\documentclass[10pt,twocolumn,letterpaper]{article}

\usepackage{cvpr}
\usepackage{times}
\usepackage{epsfig}
\usepackage{graphicx}
\usepackage{amsmath}
\usepackage{amssymb}
\usepackage{multirow}
\usepackage{subfigure}
\usepackage{algorithm}
\usepackage{algorithmic}
\usepackage{booktabs}
\usepackage{authblk}

\usepackage[pagebackref=true,breaklinks=true,letterpaper=true,colorlinks,bookmarks=false]{hyperref}

\cvprfinalcopy 

\def\cvprPaperID{1029} 

\ifcvprfinal\fi
\begin{document}

\title{HRank: Filter Pruning using High-Rank Feature Map}


\author{Mingbao Lin$^1$, Rongrong Ji$^{1,5}$\thanks{Corresponding author}\;, Yan Wang$^2$,  Yichen Zhang$^1$, \vspace{-0.8em} \\ Baochang Zhang$^3$, Yonghong Tian$^{4,5}$, Ling Shao$^6$\\
$^1$Media Analytics and Computing Laboratory, Department of Artificial Intelligence, School of Informatics, Xiamen University, China, $^2$Pinterest, USA, $^3$Beihang University, China \\ $^4$Peking University, Beijing, China, $^5$Peng Cheng Laboratory, Shenzhen, China \\ $^6$Inception Institute of Artificial Intelligence, Abu Dhabi, UAE\\
{\tt\small lmbxmu@stu.xmu.edu.cn, rrji@xmu.edu.cn, yanw@pinterest.com, ethan.zhangyc@gmail.com, bczhang@buaa.edu.cn, yhtian@pku.edu.cn, ling.shao@ieee.org}}

\maketitle
\thispagestyle{empty}

\begin{abstract}
Neural network pruning offers a promising prospect to facilitate deploying deep neural networks on resource-limited devices.
However, existing methods are still challenged by the training inefficiency and labor cost in pruning designs, due to missing theoretical guidance of non-salient network components.
In this paper, we propose a novel filter pruning method by exploring the High Rank of feature maps (HRank).
Our HRank is inspired by the discovery that the average rank of multiple feature maps generated by a  single filter is always the same, regardless of the number of image batches CNNs receive.
Based on HRank, we develop a method that is mathematically formulated to prune filters with low-rank feature maps.
The principle behind our pruning is that low-rank feature maps contain less information, and thus pruned results can be easily reproduced.
Besides, we experimentally show that weights with high-rank feature maps contain more important information, such that even when a portion is not updated, very little damage would be done to the model performance.
Without introducing any additional constraints, HRank leads to significant improvements over the state-of-the-arts in terms of FLOPs and parameters reduction, with similar accuracies.
For example, with ResNet-110, we achieve a $58.2\%$-FLOPs reduction by removing $59.2\%$ of the parameters, with only a small loss of $0.14\%$ in top-1 accuracy on CIFAR-10.
With Res-50, we achieve a $43.8\%$-FLOPs reduction by removing $36.7\%$ of the parameters, with only a loss of $1.17\%$ in the top-1 accuracy on ImageNet. The codes can be available at \url{https://github.com/lmbxmu/HRank}.
\end{abstract}

\section{Introduction}\label{introduction}
Convolutional Neural Networks (CNNs) have demonstrated great success in computer vision applications, like classification \cite{szegedy2015going,he2016deep}, detection \cite{girshick2014rich,ren2015faster}, and segmentation \cite{long2015fully,chen2017deeplab}.
However, their high demands in computing power and memory footprint prohibit most state-of-the-art CNNs to be deployed in edge devices such as smartphones or wearable devices.
While good progress has been made in designing new hardware and/or hardware-specific CNN acceleration frameworks like TensorRT, it retains as a significant demand to reduce the FLOPs and size of CNNs, with limited compromise on accuracy~\cite{cheng2018recent}.
%
%
Popular techniques include filter compression \cite{denton2014exploiting}, parameter quantization \cite{chen2015compressing}, and network pruning \cite{han2015learning}.

Among them, network pruning has shown broad prospects in various emerging applications.
Typical works either prune the filter weights to obtain sparse weight matrices (weight pruning)~\cite{han2015learning,han2015deep,carreira2018learning}, or remove entire filters from the network (filter pruning)~\cite{li2017pruning,lin2020filter,lin2020channel}.
Besides network compression, weight pruning approaches can also achieve acceleration on specialized software~\cite{park2016faster} or hardware~\cite{han2016eie}.
However, they have limited applications on general-purpose hardware or BLAS libraries.
In contrast, filter pruning approaches do not have this limitation since entire filters are removed.
In this paper, we focus on the filter pruning to achieve model compression (reduction in parameters) and acceleration (reduction in FLOPs), aiming to provide a versatile solution for devices with low computing power.

The core of filter pruning lies in the selection of filters, which should yield the highest compression ratio with the lowest compromise in accuracy.
Based on the design of filter evaluation functions, we empirically categorize filter pruning into two groups as discussed below.

\textbf{Property Importance}: The filters are pruned based on the intrinsic properties of CNNs.
These pruning approaches do not modify the network training loss.
After the pruning, the model performance is enhanced through fine-tuning.
Among these methods, Hu \emph{et al}. \cite{hu2016network} utilized the sparsity of outputs in a large network to remove filters with a high percentage of zero activations.
The ${\ell}_1$-norm based pruning \cite{li2017pruning} assumes that parameters or features with small norms are less informative, and thus should be pruned first.
Molchanov \emph{et al}. \cite{molchanov2016pruning} considered the first-order gradient to evaluate the importance of filters and removed the least important ones.
In \cite{yu2018nisp}, the importance scores of the final responses are propagated to every filter in the network and the CNNs are pruned by removing the one with the least importance.
He \emph{et al}. \cite{he2019filter} calculated the geometric median in the layers, and the filters closest to this are pruned.
Most designs for filter evaluation functions are ad-hoc, which brings the advantage of low time complexity, but also limits the acceleration and compression ratios.

\textbf{Adaptive Importance}: Different from the property importance based approaches, another direction embeds the pruning requirement into the network training loss, and employs a joint-retraining optimization to generate an adaptive pruning decision.
Liu \emph{et al}.~\cite{liu2017learning} and Zhao \emph{et al}.~\cite{zhao2019variational} imposed a sparsity constraint on the scaling factor of the batch normalization layer, allowing channels with lower scaling factors to be identified as unimportant.
Huang \emph{et al}. \cite{huang2018data} and Lin \emph{et al}. \cite{lin2019towards} introduced a new scaling factor parameter (also known as a mask) to learn a sparse structure pruning where filters corresponding to a scaling factor of zero are removed.
Compared with property importance based filter pruning, adaptive importance based methods usually yield better compression and acceleration results due to their joint optimization.
However, because the loss is changed, the required retraining step is heavy in both machine time and human labor, usually demanding another round of hyper-parameter tuning.
For some methods, \emph{e.g.}, the mask-based scheme, the modified loss even requires specialized optimizers \cite{huang2018data,lin2019towards}, which affects the flexibility and ease of using approaches based on adaptive importance.

\begin{figure}[!t]
\begin{center}
\includegraphics[height=0.4\linewidth]{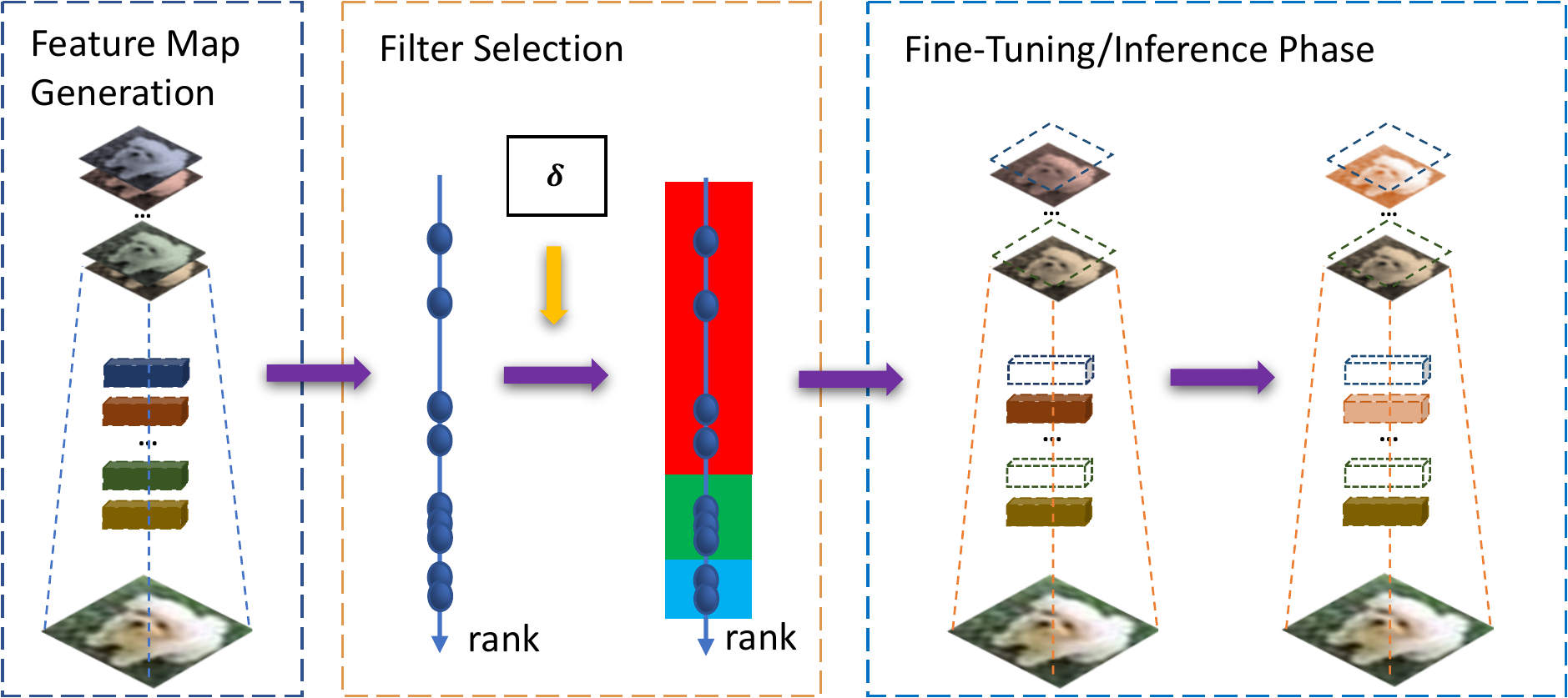}
\end{center}
\vspace{-0.8em}
\caption{\label{framework}
Framework of HRank.
In the left column, we first use images to run through the convolutional layers to get the feature maps.
In the middle column, we then estimate the rank of each feature map, which is used as the criteria for pruning.
The right column shows the pruning (the red filters), and fine-tuning where the green filters are updated and the blue filters are frozen.
}
\vspace{-1.2em}
\end{figure}

\begin{figure*}[th]
\begin{center}
\begin{minipage}[t]{0.19\linewidth}
\centerline{
\subfigure[VGGNet-16\_1.]{
\includegraphics[width=\linewidth]{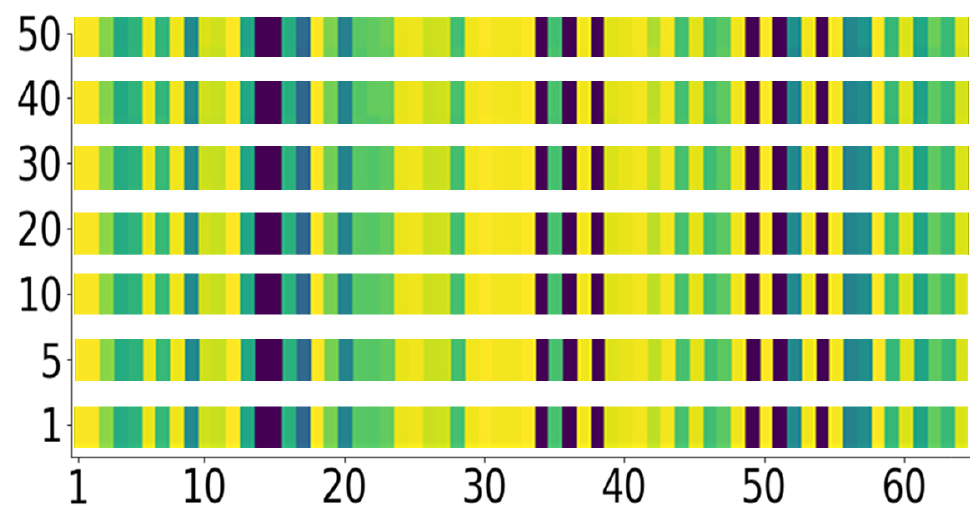}}
\subfigure[VGGNet-16\_6.]{
\includegraphics[width=\linewidth]{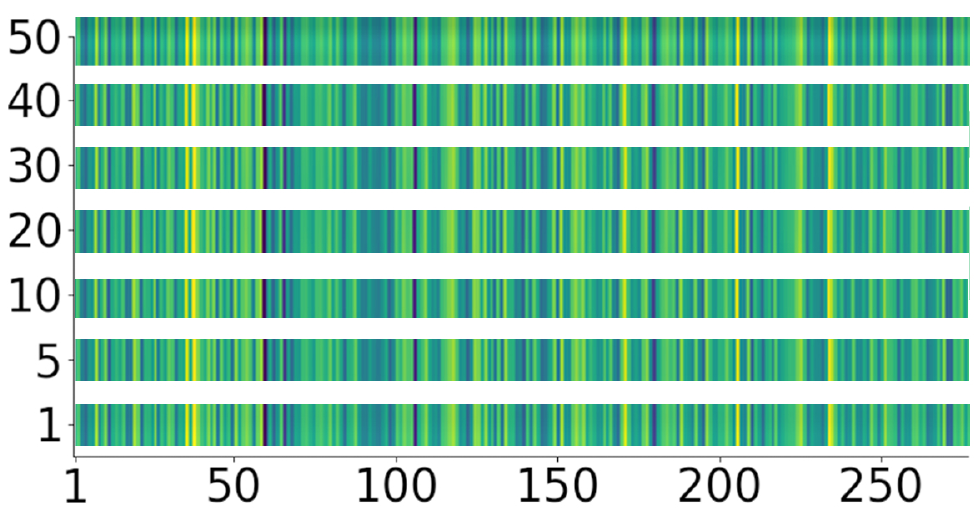}}
\subfigure[VGGNet-16\_12.]{
\includegraphics[width=\linewidth]{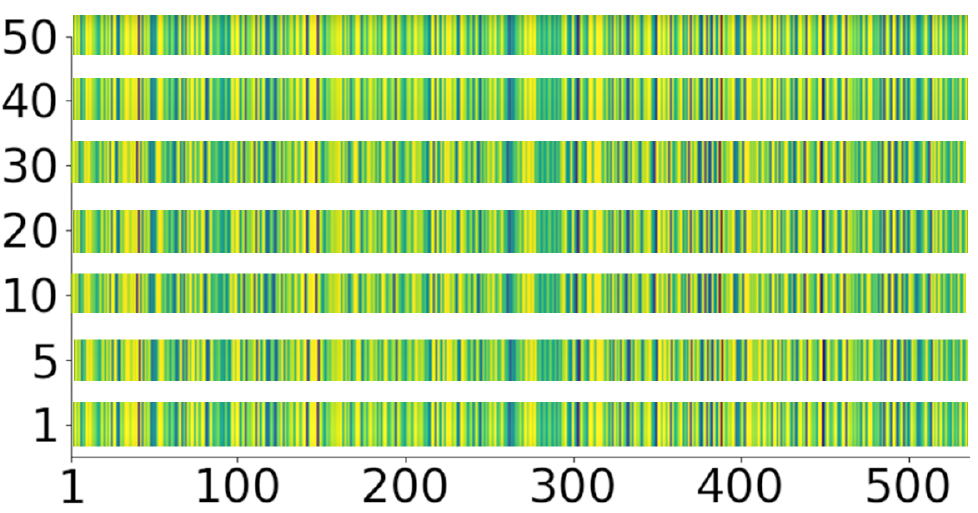}}
\subfigure[GoogLeNet\_1.]{
\includegraphics[width=\linewidth]{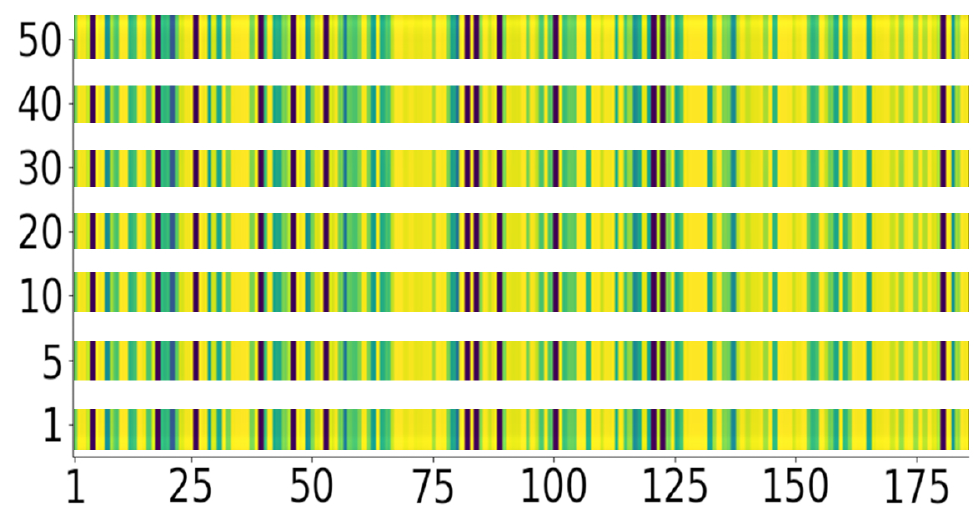}}
\subfigure[GoogLeNet\_5\_3x3.]{
\includegraphics[width=\linewidth]{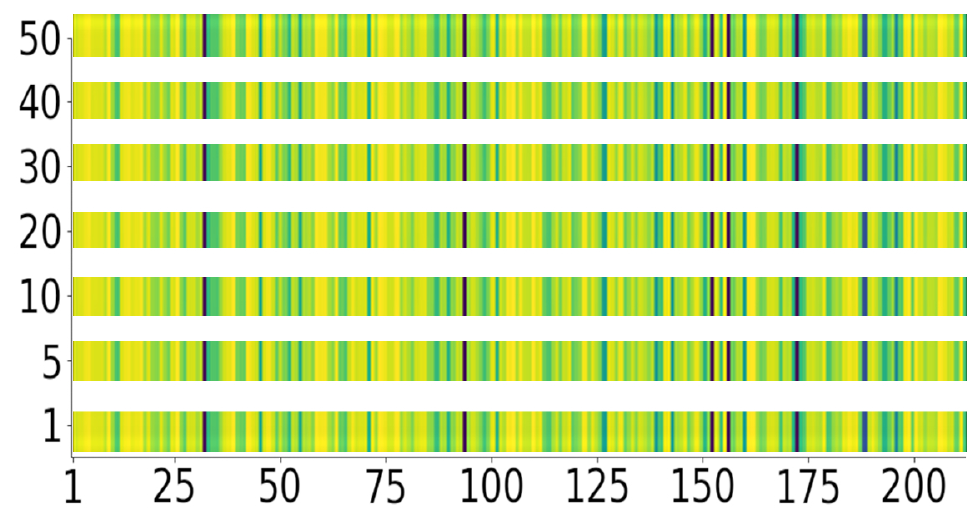}}
}
\end{minipage}

\begin{minipage}[t]{0.19\linewidth}
\centerline{
\subfigure[GoogLeNet\_10\_5x5.]{
\includegraphics[width=\linewidth]{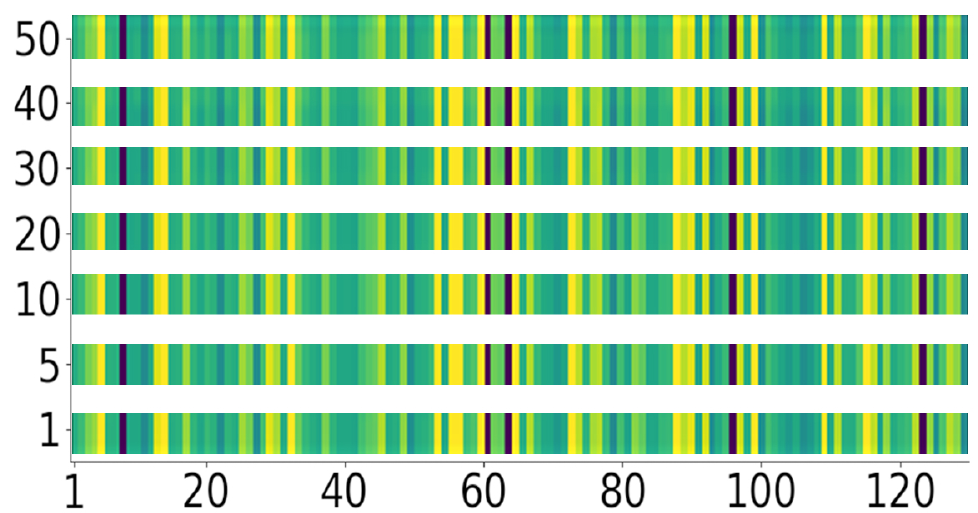}}
\subfigure[ResNet-56\_1.]{
\includegraphics[width=\linewidth]{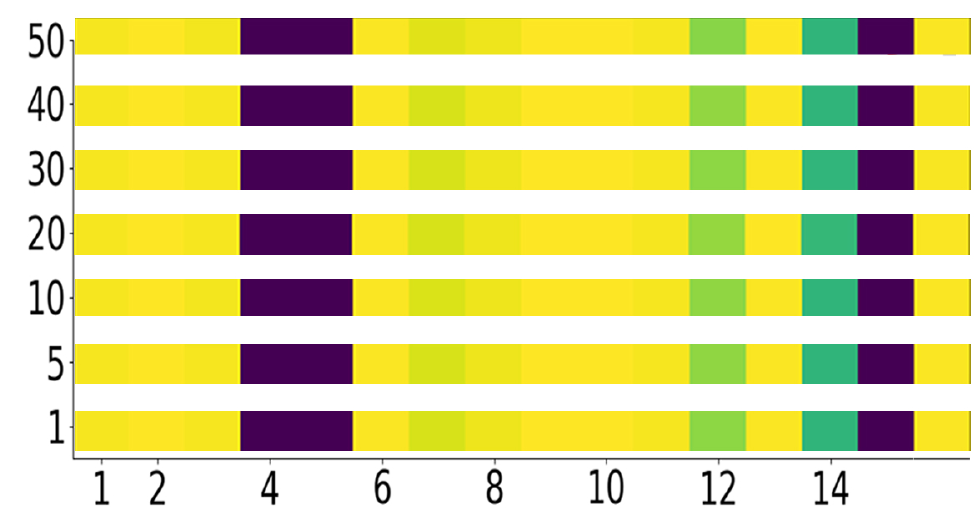}}
\subfigure[ResNet-56\_28.]{
\includegraphics[width=\linewidth]{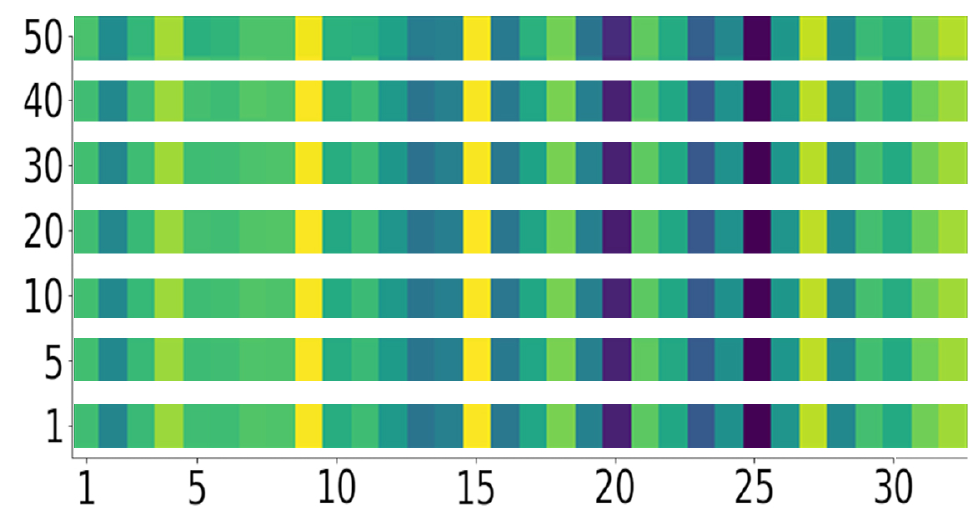}}
\subfigure[ResNet-56\_55.]{
\includegraphics[width=\linewidth]{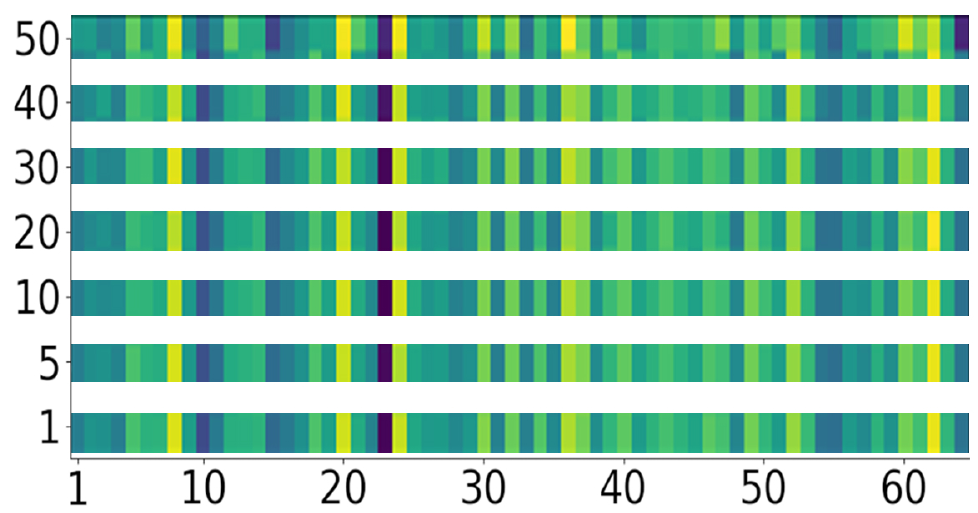}}
\subfigure[ResNet-110\_1.]{
\includegraphics[width=\linewidth]{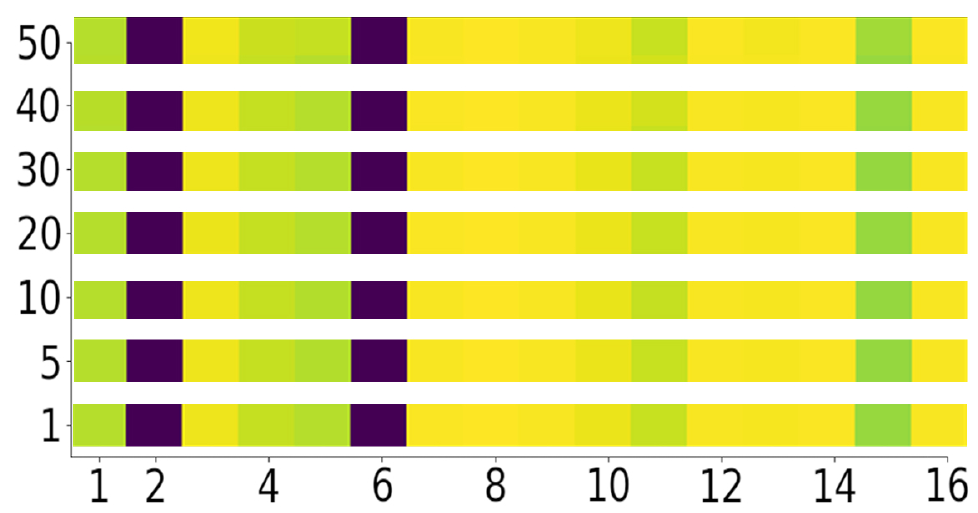}}
}
\end{minipage}

\begin{minipage}[t]{0.19\linewidth}
\centerline{
\subfigure[ResNet-110\_54.]{
\includegraphics[width=\linewidth]{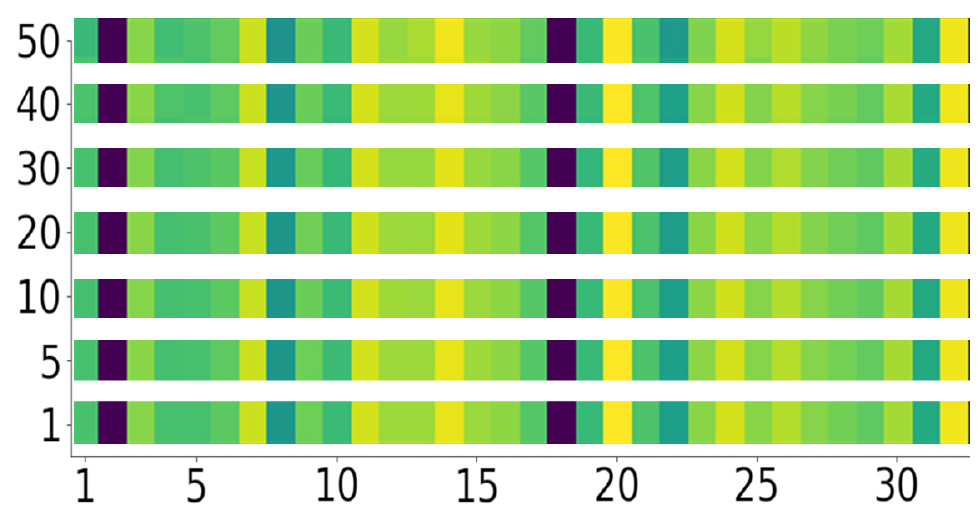}}
\subfigure[ResNet-110\_108.]{
\includegraphics[width=\linewidth]{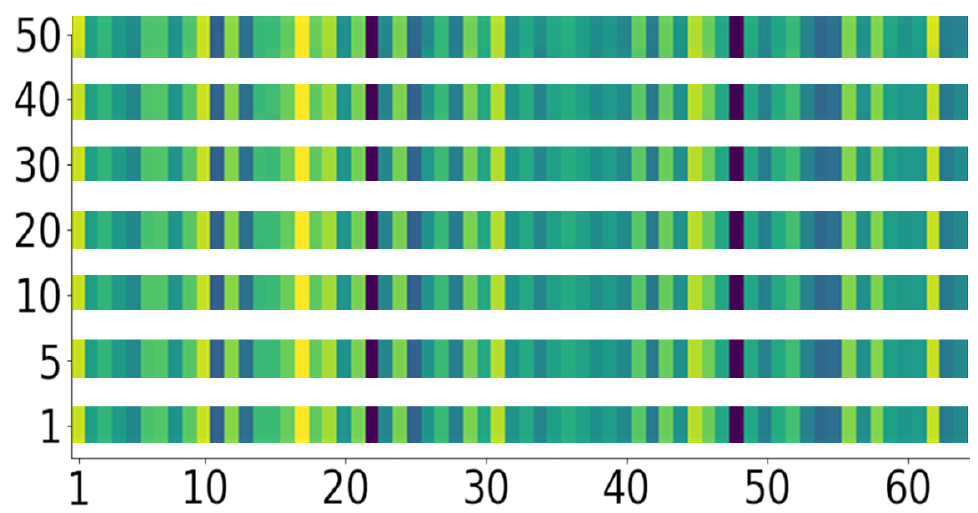}}
\subfigure[DenseNet-40\_1.]{
\includegraphics[width=\linewidth]{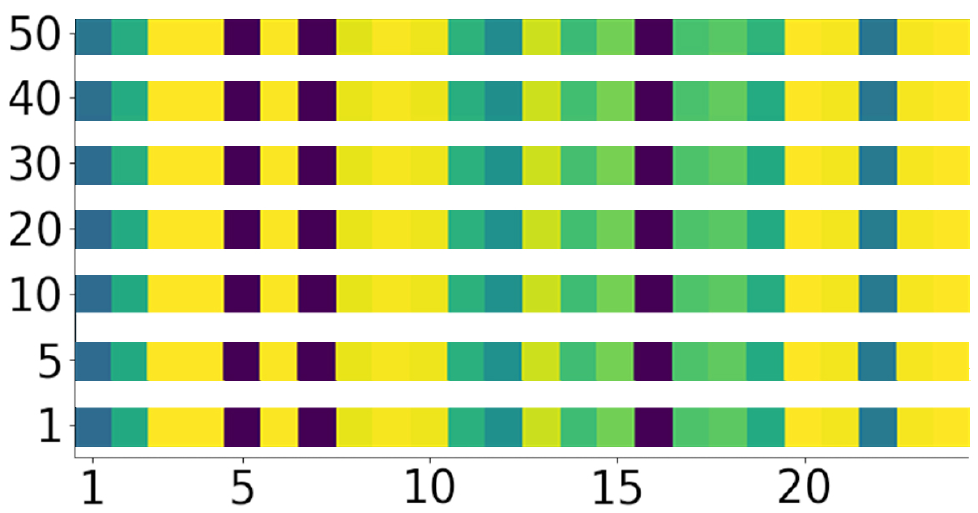}}
\subfigure[DenseNet-40\_20.]{
\includegraphics[width=\linewidth]{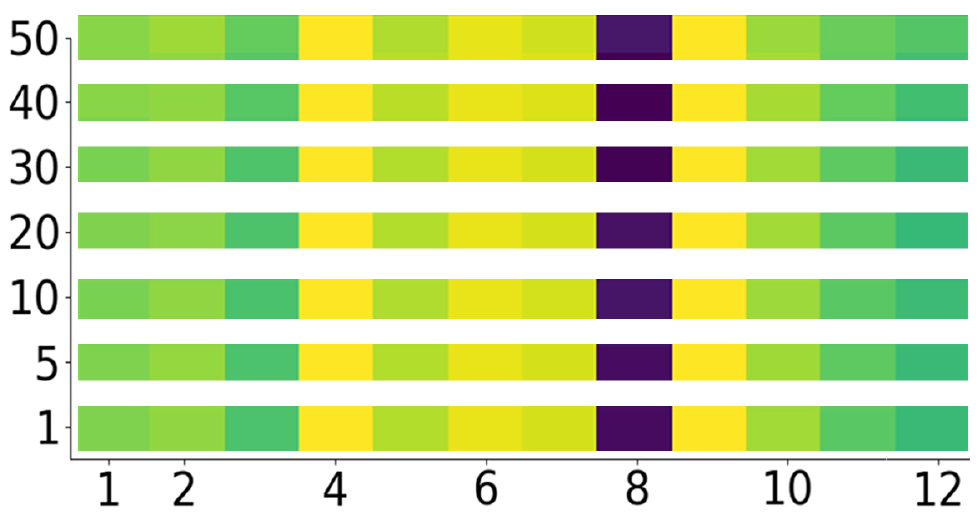}}
\subfigure[DenseNet-40\_39.]{
\includegraphics[width=\linewidth]{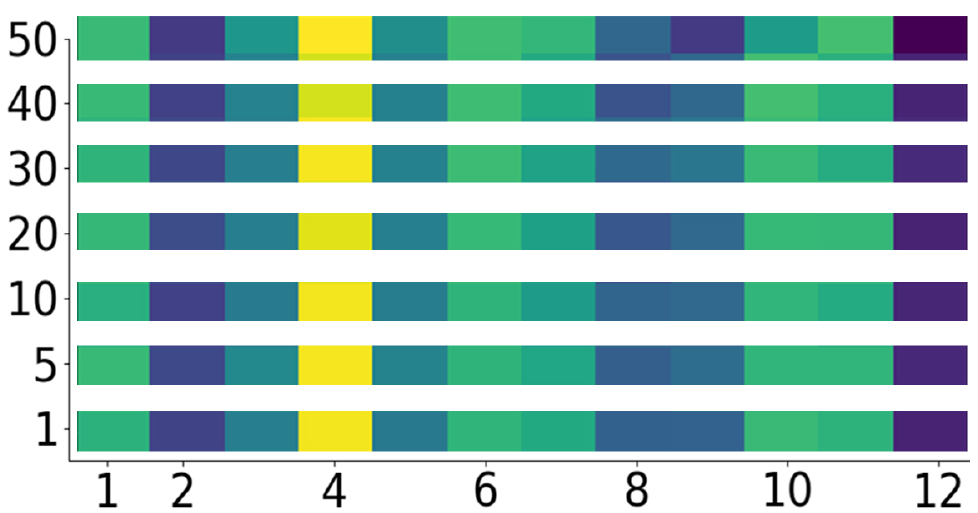}}
}
\end{minipage}

\end{center}
\vspace{-0.8em}
\caption{\label{fig_rank}Average rank statistics of feature maps from different convolutional layers and architectures on CIFAR-10.
For each subfigure, the x-axis represents the indices of feature maps and the y-axis is the batches of training images (each batch size is set to 128).
Different colors denote different rank values.
As can be seen, the rank of each feature map (column of the subfigure) is almost unchanged (the same color), regardless of the image batches.
Hence, even a small number of images can effectively estimate the average rank of each feature map in different architectures.}
\vspace{-1.2em}
\end{figure*}

%
Overall, filter pruning remains an open problem so far. On one hand we pursuit higher compression/acceleration ratios, while on the other hand we are restricted by heavy machine time and human labor (especially for adaptive importance based methods).
%
We attribute these problems to the lack of practical/theoretical guidance regarding to the filter importance and redundancy.
In this paper, we propose an effective and efficient filter pruning approach that explores the High Rank of the feature map in each layer (HRank), as shown in Fig.\,\ref{framework}.
The proposed HRank performs as such a guidance, which is a property importance based filter pruner.
It eliminates the need of introducing additional auxiliary constraints or retraining the model, thus simplifying the pruning complexity.
Moreover, compared with existing property importance based methods, HRank also leads to significant improvements in acceleration and compression \cite{li2017pruning,yu2018nisp}                             , achieving even better results than state-of-the-art adaptive importance based methods \cite{he2017channel,liu2017learning,luo2017thinet,huang2018data,lin2019towards,zhao2019variational}.
The motivation behind enforcing the high rank of feature maps comes from an empirical and quantitative observation.
As shown in Fig.~\ref{fig_rank}, we have found that the average rank of feature maps generated by a single filter is always the same, regardless of how much data the CNN has seen.
It suggests that the ranks of feature maps in deep CNNs can be accurately estimated using only a small portion of the input images, and thus can be highly efficient.
Based on this idea, we mathematically prove that feature maps with lower ranks contribute less to accuracy.
Thus the filters generating these feature maps can be removed first.

We conduct extensive experiments on two benchmarks, CIFAR-10 \cite{krizhevsky2009learning} and ImageNet \cite{russakovsky2015imagenet}, using many representative large CNN models, including VGGNet \cite{simonyan2014very}, GoogLeNet \cite{szegedy2015going}, ResNet \cite{he2016deep} and DenseNet \cite{huang2017densely}.
The results demonstrate the superior performance of HRank over existing filter pruning methods, including property importance based approaches\cite{li2017pruning,yu2018nisp}                         and state-of-the-art adaptive importance based models \cite{he2017channel,liu2017learning,luo2017thinet,huang2018data,lin2019towards,zhao2019variational}.
%
We also carry out experiments to show that, even when we freeze a portion of filters with high-rank feature maps during the fine-tuning after pruning, the model still retains high accuracy, which verifies our assumption.

To summarize, our main contributions are three-fold:
\begin{itemize}
  \item Based upon extensive statistical verification, we empirically demonstrate that the average rank of feature maps generated by a single filter is almost unchanged. To the best of our knowledge, this is the first time this observation is reported.
  \item We mathematically prove that filters with lower-rank feature maps are less informative and thus less important to preserve accuracy, which can be removed first.
  The opposite is true for high-rank feature maps.
  \item Extensive experiments demonstrate the efficiency and effectiveness of HRank in both model compression and acceleration over a variety of state-of-the-arts \cite{li2017pruning,yu2018nisp,he2017channel,liu2017learning,luo2017thinet,huang2018data,lin2019towards,zhao2019variational}.
\end{itemize}

\section{Related Work}\label{related_work}
\textbf{Filter Pruning}.
Opposed to weight pruning, which prunes the weight matrices, filter pruning removes the entire filters according to certain metrics. 
Filter pruning not only significantly reduces storage usage, but also decreases the computational cost in online inference.
As discussed in Sec.\;\ref{introduction}, filter pruning can be divided into two categories: property importance approaches and adaptive importance approaches.
Property importance based filter pruning aims to exploit the intrinsic properties of CNNs (\emph{e.g.}, ${\ell}_1$-norm and gradient, \emph{etc}), and then uses them as criteria for discriminating less important filters.
For example,
the sparser filters in \cite{molchanov2016pruning} are considered less important while \cite{hu2016network} prunes filters with smaller ${\ell}_1$-norm.
In contrast, adaptive importance based filter pruning typically retrain the networks with additional constraints to generate an adaptive pruning decision.
Beyond the related work discussed in Sec.\,\ref{introduction}, Luo \emph{et al.} \cite{luo2017thinet} established filter pruning as an optimization problem, where the filters are pruned based on the statistics information from the next layer.
He \emph{et al.} \cite{he2017channel} proposed a two-step algorithm including a LASSO regression to remove redundant filters and a linear least square to construct the outputs.

\textbf{Low-rank Decomposition}.
As shown in \cite{denil2013predicting}, neural networks tend to be over-parameterized, which suggests that the parameters in each layer can be accurately recovered from a small subset.
Inspired by this, low-rank decomposition has emerged as an alternative for network compression.
It approximates convolutional operations by representing the weight matrix as a low-rank product of two smaller matrices. 
Unlike pruning, it aims to reduce the computational costs of the network instead of changing the original number of filters.
To this end, Denton \emph{et al}. \cite{denton2014exploiting} exploited the linear structure of CNNs by exploring an appropriate low-rank approximation of the weights and keeping the accuracy within a $1\%$ loss of the original model.
Further, Zhang \emph{et al}. \cite{zhang2015efficient} considered the subsequent nonlinear units while learning the low-rank decomposition to provide significant speed-up.
Lin \emph{et al}. \cite{lin2018holistic} proposed a low-rank decomposition for both convolutional filters and fully-connected matrices, which are then solved by a closed-form solver with a fixed rank.
Though low-rank decomposition can benefit the compression and speedup of CNNs, it normally incurs a large loss in accuracy under high compression ratios.

\textbf{Discussion}.
Compared with weight pruning, filter pruning 
tends to be more favorable in reducing model complexity.
Besides, its structured pruning can be easily integrated into the highly efficient BLAS libraries without specialized software or hardware support.
Nevertheless, existing filter pruning methods suffer from inefficient acceleration and compression (property importance based filter pruning) or machine and labor costs (adaptive importance based filter pruning), as discussed in Sec.\;\ref{introduction}.
These two issues bring fundamental challenges to the deployment of deep CNNs on resource-limited devices.
%
We attribute the dilemma to the missing of practical/theoretical guidance regarding to the filter importance and redundancy.
We focus on the rank of feature maps and analyze its effectiveness both theoretically and experimentally.
Note that our approach is orthogonal to low-rank decomposition approaches.
We aim to prune filters generating low-rank feature maps rather than decompose filters.
Note that, the low-rank methods can be integrated into our model (\emph{e.g.}, to decompose fully-connected layers) to achieve higher compression and speedup ratios.

\section{The Proposed Method}\label{the_proposed_method}
\subsection{Notations}\label{terminology}
Assume a pre-trained CNN model has a set of $K$ convolutional layers%
\footnote{For simplicity, a convolutional layer includes the non-linear operations, \emph{e.g.}, pooling, batch normalization, ReLU, dropout and so on.
Besides, for ease of FLOPs computation and parameters calculation, in this paper, we ignore the cost of these non-linear operations.}, and ${\mathcal{C}}^i$ is the $i$-th convolutional layer.
The parameters in $\mathcal{C}^i$ can be represented as a set of $3$-D filters $\mathcal{W}_{\mathcal{C}^i} = \{ \mathbf{w}_1^i, \mathbf{w}_2^i, ..., \mathbf{w}_{n_i}^i \} \in \mathbb{R}^{n_{i} \times n_{i-1} \times k_i \times k_i}$, where the $j$-th filter is $\mathbf{w}_j^i \in \mathbb{R}^{n_{i-1} \times k_i \times k_i}$.
$n_i$ represents the number of filters in $\mathcal{C}^i$ and $k_i$ denotes the kernel size.
The outputs of filters, \emph{i.e.}, feature maps, are denoted as $\mathcal{O}^i = \{ \mathbf{o}_1^i, \mathbf{o}_2^i, ..., \mathbf{o}_{n_i}^i \} \in \mathbb{R}^{n_i \times g \times h_i \times w_i}$, where the $j$-th feature map $\mathbf{o}_j^i \in
\mathbb{R}^{g \times h_i \times w_i}$ is generated by $\mathbf{w}_j^i$.
$g$ is the size of input images.
$h_i$ and $w_i$ are the height and width of the feature map, respectively.
In filter pruning, $\mathcal{W}_{C^i}$ can be split into two groups, \emph{i.e.}, a subset to be kept $\mathcal{I}_{\mathcal{C}^i} = \{ \mathbf{w}^i_{\mathcal{I}^i_1}, \mathbf{w}^i_{\mathcal{I}^i_2}, ..., \mathbf{w}^i_{\mathcal{I}^i_{n_{i1}}}\}$ and a subset, with less importance, to be pruned $\mathcal{U}_{\mathcal{C}^i} = \{ \mathbf{w}^i_{\mathcal{U}^i_1}, \mathbf{w}^i_{\mathcal{U}^i_2}, ..., \mathbf{w}^i_{\mathcal{U}^i_{n_{i2}}}\}$,
where $\mathcal{I}_j^i$ and $\mathcal{U}_j^i$ are the indices of the $j$-th important and unimportant filter, respectively.
$n_{i1}$ and $n_{i2}$ are the number of important and unimportant filters, respectively.
We have:
$\mathcal{I}_{{\mathcal{C}}^i} \cap \mathcal{U}_{{\mathcal{C}}^i} = \varnothing$, $\mathcal{I}_{{\mathcal{C}}^i} \cup \mathcal{U}_{{\mathcal{C}}^i} = \mathcal{W}_{\mathcal{C}^i}$ and $n_{i1} + n_{i2} = n_i$.

\subsection{HRank}\label{the_proposed_approach}
Filter pruning aims to identify and remove the less important filter set from $\mathcal{I}_{\mathcal{C}^i}$, which can be formulated as an optimization problem:
\begin{equation}\label{filter_importance}
\begin{split}
&\min\limits_{{\delta}_{ij}}\sum_{i=1}^K\sum_{j=1}^{n_i} {\delta}_{ij}\mathcal{L}(\mathbf{w}_j^i),
\\&
s.t. \; \sum_{j=1}^{n_i}{\delta}_{ij} = n_{i2},
\end{split}
\end{equation}
where ${\delta}_{ij}$ is an indicator which is $1$ if $\mathbf{w}_j^i$ is grouped to $\mathcal{U}_{C^i}$, or $0$ if $\mathbf{w}_j^i$ is grouped to $\mathcal{I}_{C^i}$.
$\mathcal{L}(\cdot)$ measures the importance of a filter input to the CNN.
Thus minimizing Eq.\,(\ref{filter_importance}) is equal to removing the $n_{i2}$ least important filters in $\mathcal{C}^i$.

In this framework, most prior works resort to directly designing $\mathcal{L}(\cdot)$ on the filters.
We argue that ad-hoc designs of $\mathcal{L}$ on the filters ignore the distribution of the input images as well as the output labels, and might be the reason why property importance based approaches show suboptimal performance.
In contrast, in this paper, we propose to define $\mathcal{L}$ on the feature maps.
The rationale lies in that the feature maps are an intermediate step that can reflect both the filter properties and the input images.
As investigated in \cite{zhou2018revisiting}, individual feature maps, even within the same layer, play different roles in the network.
Furthermore, they can capture how an input image is transformed in each layer, and finally, to the predicted labels.
To that effect, we reformulate Eq.\,(\ref{filter_importance}) as:

\begin{equation}\label{feature_map_importance}
\begin{split}
&\min\limits_{{\delta}_{ij}}\sum_{i=1}^K\sum_{j=1}^{n_i} {\delta}_{ij}\mathbb{E}_{I \sim P(I)}\big[\hat{\mathcal{L}}\big(\mathbf{o}_j^i(I,:,:)\big)\big],
\\& \qquad\quad
s.t. \; \sum_{j=1}^{n_i}{\delta}_{ij} = n_{i2},
\end{split}
\end{equation}
where $I$ is the input image, which is sampled from a distribution $P(I)$.
$\hat{\mathcal{L}}(\cdot)$ is used to estimate the information of feature maps $\mathbf{o}_j^i(I,:,:)$ generated by the filter $\mathbf{w}_j^i$.
In our settings, the expectation of $\hat{\mathcal{L}}(\cdot)$ is proportional to $\mathcal{L}(\cdot)$ in Eq.\,(\ref{filter_importance}), \emph{i.e.}, the more information the feature map contains, the more important the corresponding filter is.

Thus our key problem falls in designing a proper function $\hat{\mathcal{L}}(\cdot)$ which can well reflect the information richness of feature maps.
%
Prior property importance based method \cite{hu2016network} utilizes the sparsity of feature maps.
It calculates the zero activations from extensive input images and removes the filters with a large portion of zeros in the feature maps.
However, such a data-driven scheme heavily relies on the distribution $P(I)$ of the input images, which is extremely complex to describe.
While the Monte-Carlo approximation can be adopted, it is still intractable since a large number of input images (over ten thousand in \cite{hu2016network}) are required to maintain a reasonable prediction.

Instead, we exploit the rank of feature maps which is demonstrated to be not only an effective measure of information, but also a stable representation across $P(I)$.
We first define our information measurement as:
\begin{equation}\label{feature_map_rank}
\hat{L}\big(\mathbf{o}^i_j(I,:,:)\big) = \textbf{Rank}\big(\mathbf{o}_j^i(I,:,:)\big),
\end{equation}
where $\textbf{Rank}(\cdot)$ is the rank of a feature map for input image $I$.
%
We conduct a Singular Value Decomposition (SVD) for $\mathbf{o}^i_j(I,:,:)$:
\begin{equation}\label{decomposition}
\begin{split}
\mathbf{o}_j^i(I,:,:) & = \sum_{i=1}^{r}{\sigma}_i\mathbf{u}_i\mathbf{v}_i^T
\\&
= \sum_{i=1}^{r'}{\sigma}_i\mathbf{u}_i\mathbf{v}_i^T + \sum_{i=r' + 1}^{r}{\sigma}_i\mathbf{u}_i\mathbf{v}_i^T,
\end{split}
\end{equation}
where $r = \textbf{Rank}(\mathbf{o}_j^i(I,:,:))$ and $r' < r$.
${\delta}_i$, $\mathbf{u}_i$ and $\mathbf{v}_i$ are the top-$i$ singular values, left singular vector and right singular vector of $\mathbf{o}_j^i(I,:,:)$, respectively.
It can be seen that a feature map with rank $r$ can be decomposed into a lower-rank feature map with rank $r'$, \emph{i.e.}, $\sum_{i=1}^{r'}{\sigma}_i\mathbf{u}_i\mathbf{v}_i^T$, and some additional information, \emph{i.e.}, $\sum_{i=r' + 1}^{r}{\sigma}_i\mathbf{u}_i\mathbf{v}_i^T$.
Hence, higher-rank feature maps actually contain more information than lower-rank ones.
Thus, the rank can serve as a reliable measurement for information richness.

\subsection{Tractability of Optimization}
One critical point in the pruning procedures discussed above falls in the calculation of ranks.
However, it is possible that the rank of the feature maps $\mathbf{o}^i_j \in \mathbb{R}^{g \times h_i \times w_i}$ is sensitive to the distribution of input images, $P(I)$.
For example, the rank of $\mathbf{o}^i_j(1,:,:)$ is not necessarily the same as that of $\mathbf{o}^i_j(2,:,:)$.
To accurately estimate the expectation of ranks, extensive input images have to be used, as with \cite{hu2016network}.
This poses a great challenge when evaluating the relative importance of filters.

Fortunately, we empirically observe that the expectation of ranks generated by a single filter is robust to the input images.
To illustrate, we plot the average of rank values \emph{w.r.t.} different numbers of input batches in Fig.\;\ref{fig_rank}.
It is obvious that the colors (which indicate the values) of the average ranks from a single filter are the same, which are irrespective to the image that CNNs receive.
To explain, although different images may have different ranks, the variance is negligible.
Hence, a small batch of input images can be used to accurately estimate the expectation of the feature map rank.
Then, we define the rank expectation of feature map $\mathbf{o}_j^i$ as:
\begin{equation}\label{rank_expectation}
\mathbb{E}_{I \sim P(I)}\big[\hat{\mathcal{L}}\big(\mathbf{o}_j^i(I,:,:)\big)\big] \approx \sum_{t=1}^g\textbf{Rank}\big(\mathbf{o}_j^i(t,:,:)\big).
\end{equation}

In our experiments, we set the number of images, $g = 500$, to estimate the average rank, which is several orders of magnitude smaller than \cite{hu2016network}, where over ten thousand input images are used to estimate the sparsity of feature maps.
Besides, calculating the ranks of feature maps can be implemented offline.
Hence, Eq.\,(\ref{feature_map_rank}) is easily tractable using the above approximation.
Combining Eq.\,(\ref{feature_map_importance}) and Eq.\,(\ref{rank_expectation}), we derive our final optimization objective as follows:

\begin{equation}\label{rank_importance}
\begin{split}
\min\limits_{{\delta}_{ij}} \sum_{i=1}^K\sum_{j=1}^{n_i} &{\delta}_{ij}(\mathbf{w}_j^i)\sum_{t=1}^g\textbf{Rank}\big(\mathbf{o}_j^i(t,:,:)\big),
\\& s.t. \; \sum_{j=1}^{n_i}{\delta}_{ij} = n_{i2}.
\end{split}
\end{equation}
It is intuitive that Eq.\,(\ref{rank_importance}) can be minimized by pruning the filters with $n_{i2}$ least average ranks of feature maps.

\textbf{Pruning Procedure}.
Given filters $\mathcal{W}_{\mathcal{C}^i}$ in $\mathcal{C}^i$ and their generated feature maps $\mathcal{O}^i$, as defined in Sec.\,\ref{terminology}, we prune $\mathcal{W}_{\mathcal{C}^i}$ as follows:
First, we calculate the average rank of feature map $\mathbf{o}^i_j$ in $\mathcal{O}^i$, forming a rank set $\mathcal{R}^i = \{ r^i_1, r^i_2, ..., r^i_{n_i} \} \in \mathbb{R}^{n_i}$.
Second, we re-rank the rank set in decreasing order $\hat{\mathcal{R}}^i = \{ r^i_{I^i_1}, r^i_{I^i_2}, ..., r^i_{I^i_{n_i}} \} \in \mathbb{R}^{n_i}$, where $I^i_j$ is the index of the $j$-th top value in $\mathcal{R}^i$.
Third, we determine the values of $n_{i1}$ (number of preserved filters) and $n_{i2}$ (number of pruned filters).
Fourth, we obtain the important filter set $\mathcal{I}_{\mathcal{C}^i} = \{ \mathbf{w}^i_{\mathcal{I}^i_1}, \mathbf{w}^i_{\mathcal{I}^i_2}, ..., \mathbf{w}^i_{\mathcal{I}^i_{n_{i1}}}\}$ where the rank of $\mathbf{w}^i_{\mathcal{I}^i_j}$ is $r^i_{I^i_j}$.
%
Similarly, we obtain the filter set $\mathcal{U}_{\mathcal{C}^i} = \{ \mathbf{w}^i_{\mathcal{U}^i_1}, \mathbf{w}^i_{\mathcal{U}^i_2}, ..., \mathbf{w}^i_{\mathcal{U}^i_{n_{i2}}}\}$.
Lastly, we remove the set $\mathcal{U}_{\mathcal{C}^i}$ and fine-tune the network with $\mathcal{I}_{\mathcal{C}^i}$ as the initialization.

\section{Experiments} \label{experiment}
\subsection{Experimental Settings}

\begin{figure}[!t]
\begin{center}
\includegraphics[height=0.95\linewidth]{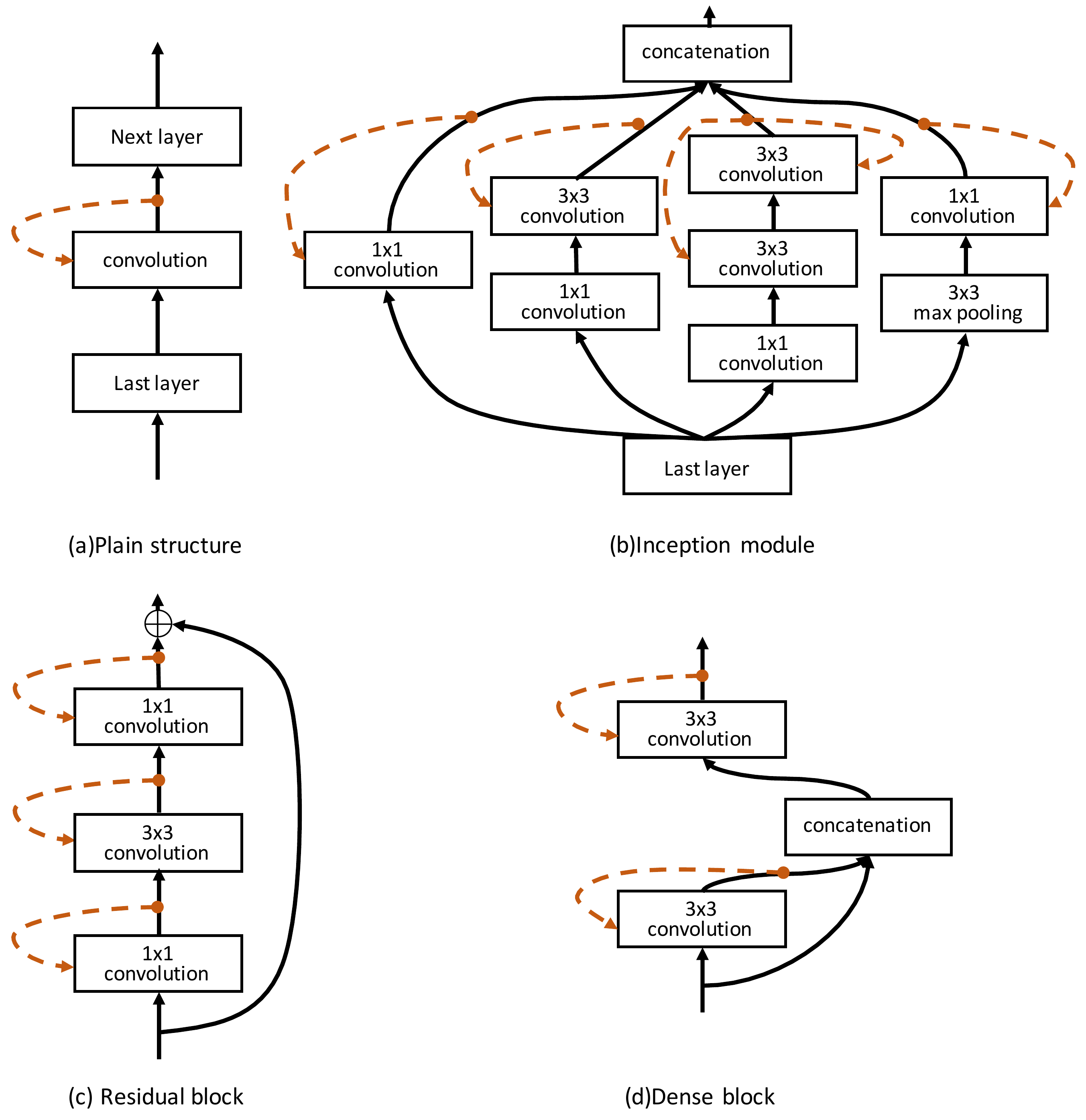}
\end{center}
\vspace{-0.8em}
\caption{\label{pruning_structure}
An illustration of mainstream network structures to be pruned, including Plain structure \cite{simonyan2014very}, Inception module \cite{szegedy2015going}, Residual block \cite{he2016deep} and Dense block \cite{huang2017densely}.
The black lines denote the inference streams of CNNs.
The red dots denote the outputs of a convolutional layer (\emph{i.e.}, feature maps).
The red dashed lines denote the to-be-pruned layers after observing the ranks of feature maps.
Note that, for $1 \times 1$ convolutions followed by $n \times n$ convolution ($n = 3$ in (b)), we do not consider pruning the $1 \times 1$ filter since it contains fewer parameters and less computation compared with an $n \times n$ filter.
\vspace{-1.2em}
}

\end{figure}

\textbf{Datasets and Baselines}.
To demonstrate our efficiency in reducing model complexity, we conduct experiments on both small and large datasets, \emph{i.e.}, CIFAR-10 \cite{krizhevsky2009learning} and ImageNet \cite{russakovsky2015imagenet}.
We study the performance of different algorithms on the mainstream CNN models, including VGGNet with a plain structure \cite{simonyan2014very}, GoogLeNet with an inception module \cite{szegedy2015going}, ResNet with a residual block \cite{he2016deep} and DenseNet with a dense block \cite{huang2017densely}.
Fig.\;\ref{pruning_structure} provides standards for pruning these popular networks.
For all benchmarks and architectures, we randomly sample $500$ images to estimate the average rank of each feature map.

\textbf{Evaluation Protocols.}
We adopt the widely-used protocols, \emph{i.e.}, number of parameters and required Float Points Operations (denoted as FLOPs), to evaluate model size and computational requirement.
%
%
To evaluate the task-specific capabilities, we provide top-1 accuracy of pruned models and the pruning rate (denoted as PR) on CIFAR-10, and top-1 and top-5 accuracies of pruned models on ImageNet.

\textbf{Configurations.}
We use PyTorch \cite{paszke2017automatic} to implement the proposed HRank approach.
We solve the optimization problem by using Stochastic Gradient Descent algorithm (SGD) with an initial learning rate of $0.01$.
The batch size, weight decay and momentum are set to $128$, $0.0005$ and $0.9$, respectively.
For each layer, we retrain the network for $30$ epochs after pruning\footnote{Better accuracy can be observed when training more epochs.
However, it requires more training time.}, with the learning rate being divided by $10$ at epochs $5$ and $10$ on CIFAR-10, and divided by $10$ every $10$ epochs on ImageNet.
All experiments are conducted on two NVIDIA GTX 1080Ti GPUs.
It is worth noting that for ResNet-50 on ImageNet, instead of fine-tuning the network every time after pruning a convolution layer, we prune the network in a per-block fashion, i.e. we fine-tune the network only after performing pruning within the entire block, as shown in Fig.\;\ref{pruning_structure} (c).
So the computational cost can be further reduced, which is important for large-scale datasets such as ImageNet.
For a fair comparison, we fix the accuracy to be similar to the baselines and measure the parameters and FLOPs reductions, or fix similar reductions of parameters and FLOPs, and measure the accuracy.

\subsection{Results and Analysis}\label{results_and_analysis}

\subsubsection{Results on CIFAR-10}\label{results_on_cifar10}
We analyze the performance on CIFAR-10, comparing against several popular CNNs, including VGG-16, GoogLeNet, ResNet56/110 and DenseNet-40.
We use a variation of VGG-16, as in \cite{li2017pruning}.
Following the state-of-the-art baseline \cite{lin2019towards}, the output of the original GoogLeNet is changed to fit the class number in CIFAR-10.
And for DenseNet-40, it has 40 layers with a growth rate of 12.

\begin{table}[]
\scriptsize
\centering
\caption{Pruning results of VGGNet on CIFAR-10.}
\label{vggnet_cifar10}
\begin{tabular}{cccc}
\toprule
Model                                                 &Top-1\%           &FLOPs(PR)        &Parameters(PR)\\
\midrule
VGGNet                                                &93.96             &313.73M(0.0\%)    &14.98M(0.0\%)     \\
L1 \cite{li2017pruning}                               &93.40             &206.00M(34.3\%)    &5.40M(64.0\%) \\
%
%
SSS \cite{huang2018data}                             &93.02             &183.13M(41.6\%)    &3.93M(73.8\%) \\
Zhao \emph{et al.} \cite{zhao2019variational}         &93.18             &190.00M(39.1\%)    &3.92M(73.3\%) \\
\textbf{HRank}(Ours)                                  &93.43             &145.61M(53.5\%)    &2.51M(82.9\%)   \\
GAL-0.05 \cite{lin2019towards}                        &92.03             &189.49M(39.6\%)    &3.36M(77.6\%)  \\
\textbf{HRank}(Ours)                                  &92.34             &108.61M(65.3\%)    &2.64M(82.1\%)   \\
GAL-0.1 \cite{lin2019towards}                         &90.73             &171.89M(45.2\%)    &2.67M(82.2\%)\\
\textbf{HRank}(Ours)                                  &91.23             &73.70M(76.5\%)    &1.78M(92.0\%)   \\
%
\bottomrule
\end{tabular}
\vspace{-1.0em}
\end{table}

\begin{table}[]
\scriptsize
\centering
\caption{Pruning results of GoogLeNet on CIFAR-10.}
\label{googlenet_cifar10}
\begin{tabular}{cccc}
\toprule
Model                                                  &Top-1\%           &FLOPs(PR)       &Parameters(PR)\\
\midrule
GoogLeNet                                              &95.05             &1.52B(0.0\%)    &6.15M(0.0\%)     \\
Random                                                 &94.54             &0.96B(36.8\%)    &3.58M(41.8\%)      \\
L1 \cite{li2017pruning}                               &94.54             &1.02B(32.9\%)    &3.51M(42.9\%) \\
\textbf{HRank}(Ours)                                   &94.53             &0.69B(54.9\%)    &2.74M(55.4\%)   \\
GAL-ApoZ \cite{hu2016network}                         &92.11             &0.76B(50.0\%)    &2.85M(53.7\%)  \\
GAL-0.05 \cite{lin2019towards}                         &93.93             &0.94B(38.2\%)    &3.12M(49.3\%)  \\
\textbf{HRank}(Ours)                                   &94.07             &0.45B(70.4\%)    &1.86M(69.8\%)   \\
\bottomrule
\end{tabular}
\vspace{-1.5em}
\end{table}

\textbf{VGG-16}.
Tab.\;\ref{vggnet_cifar10} shows the performance of different methods, including a property importance based method, \emph{i.e.}, L1 \cite{li2017pruning} and several adaptive importance based methods, \emph{i.e.}, SSS \cite{huang2018data}, Zhao \emph{et al}. \cite{zhao2019variational} and GAL \cite{lin2019towards}.
%
%
Compared with L1, HRank provides significantly better parameters and FLOPs  reductions (53.5\% \emph{vs}. 34.3\% and 82.9\% \emph{vs}. 64.0\%), which demonstrates the superiority of exploiting the rank of feature maps as an intrinsic property of CNNs.
Compared with adaptive importance based methods,
HRank yields a better compression and acceleration rate than SSS and Zhao \emph{et al}., while maintaining a higher accuracy (93.43\% \emph{vs}. 93.02\% by SSS and 93.18\% by Zhao \emph{et al}.).
Compared with GAL-0.05, HRank is advantageous in all aspects (65.3\% \emph{vs}. 39.6\% in FLOPs reduction, 82.1\% \emph{vs}. 77.6\% in parameters reduction, and 92.34\% \emph{vs}. 92.03\% in top-1 accuracy).
Lastly, compared with GAL-0.1, where only a 90.73\% top-1 accuracy is obtained, the proposed HRank gains a better result of 91.23\%.
Moreover, over $76\%$ of FLOPs are reduced and $90\%$ of parameters are removed.
Hence, HRank demonstrates its ability to compress and accelerate a neural network with a plain structure.

\textbf{GoogLeNet}.
The results for GoogLeNet are shown in Tab.\;\ref{googlenet_cifar10}.
HRank again demonstrates its ability to obtain a high accuracy of $94.53\%$, with around $55\%$ FLOPs and parameters reduction.
This is significantly better than L1 and randomized pruning, which indicates that pruning with HRank is more compact and efficient.
Furthermore, compared with adaptive importance based methods, GAL-ApoZ and GAL-0.05,
HRank obtains a better top-1 accuracy (94.07\% \emph{vs}. 92.11\% by GAL-ApoZ and 93.93\% by GAL-0.05), faster acceleration (70.4\% \emph{vs}. 50.0\% by GAL-ApoZ and 38.2\% by GAL-0.05.) and more compact parameters (69.8\% reduction \emph{vs}. 53.7\% by GAL-ApoZ and 49.3\% by GAL-0.05\%).
Without introducing additional constraints, HRank still reduces model complexity while maintaining a good accuracy, showing that the ranks of feature maps can serve as a discriminative property for identifying the redundant filters.
Besides, it demonstrates that HRank can be effectively applied to compressing neural networks with inception modules.

%
\begin{table}[]
\scriptsize
\centering
\caption{Pruning results of ResNet-56/110 on CIFAR-10.}
\label{resnet_cifar10}
\begin{tabular}{cccc}
\toprule
Model                                                  &Top-1\%           &FLOPs(PR)       &Parameters(PR)\\
\midrule
ResNet-56                                              &93.26              &125.49M(0.0\%)     &0.85M(0.0\%)     \\
L1 \cite{li2017pruning}                               &93.06              &90.90M(27.6\%)   &0.73M(14.1\%) \\
\textbf{HRank}(Ours) &93.52  &88.72M(29.3\%)  &0.71M(16.8\%) \\
NISP \cite{yu2018nisp}                                 &93.01              &81.00M(35.5\%)     &0.49M(42.4\%)  \\
GAL-0.6                                                &92.98              &78.30M(37.6\%)   &0.75M(11.8\%)  \\
\textbf{HRank}(Ours)                                   &93.17              &62.72M(50.0\%)  &0.49M(42.4\%)   \\
He \emph{et al}. \cite{he2017channel}                  &90.80              &62.00M(50.6\%)     &-      \\
GAL-0.8                                                &90.36              &49.99M(60.2\%)   &0.29M(65.9\%)  \\
\textbf{HRank}(Ours)                                   &90.72              &32.52M(74.1\%)   &0.27M(68.1\%)  \\
\midrule
ResNet-110                                             &93.50              &252.89M(0.0\%) &1.72M(0.0\%)   \\
L1 \cite{li2017pruning}                                &93.30             &155.00M(38.7\%)    &1.16M(32.6\%) \\
\textbf{HRank}(Ours) &94.23  &148.70M(41.2\%) &1.04M(39.4\%) \\
GAL-0.5 \cite{lin2019towards}                          &92.55             &130.20M(48.5\%)  &0.95M(44.8\%) \\
\textbf{HRank}(Ours)                                   &93.36             &105.70M(58.2\%)  &0.70M(59.2\%)
\\
\textbf{HRank}(Ours)                                   &92.65             &79.30M(68.6\%)  &0.53M(68.7\%)  \\
\bottomrule
\end{tabular}
\vspace{-0.5em}
\end{table}

\begin{table}[]
\scriptsize
\centering
\caption{Pruning results of DenseNet-40 on CIFAR-10.}
\label{densenet_cifar10}
\begin{tabular}{cccc}
\toprule
Model                                                  &Top-1\%            &FLOPs(PR)       &Parameters(PR)\\
\midrule
DenseNet-40                                            &94.81              &282.00M(0.0\%)    &1.04M(0.0\%)     \\
Liu \emph{et al.}-40\% \cite{liu2017learning}          &94.81              &190.00M(32.8\%)    &0.66M(36.5\%)      \\
%
%
GAL-0.01 \cite{lin2019towards}                         &94.29              &182.92M(35.3\%) &0.67M(35.6\%)  \\
\textbf{HRank}(Ours)                                   &94.24              &167.41M(40.8\%)  &0.66M(36.5\%) \\
Zhao \emph{et al.} \cite{zhao2019variational}          &93.16              &156.00M(44.8\%)&0.42M(59.7\%) \\
GAL-0.05 \cite{lin2019towards}                         &93.53              &128.11M(54.7\%) &0.45M(56.7\%)  \\
\textbf{HRank}(Ours) &93.68 &110.15M(61.0\%) &0.48M(53.8\%) \\
%
%
\bottomrule
\end{tabular}
\vspace{-1.5em}
\end{table}

\textbf{ResNet-56/110}.
%
%
We compress ResNet-56 and ResNet-110, results of which are displayed in Tab.\;\ref{resnet_cifar10}.
We begin with ResNet-56.
Under similar FLOPs and parameters reduction with L1, HRank obtains an excellent top-1 accuracy (93.52\% \emph{vs}. 93.06\%), which is even better than the baseline model (93.52\% \emph{vs}. 93.26\%).
Besides, we observe that HRank shares the same parameters reduction as NISP, another property importance based method, but achieves a better accuracy (93.17\% \emph{vs}. 93.01) and larger FLOPs reduction (50.0\% \emph{vs}. 35.5\%).
%
%
Moreover, HRank can effectively deal with filters with more computation costs, leading to a significant acceleration.
Similar observations can be found when compared with adaptive importance based methods, including He \emph{et al}. and GAL-0.8.
HRank yields an impressive acceleration (74.1\% \emph{vs}. 50.6\% for He \emph{et al}. and 60.2\% for GAL-0.8).

Next, we analyze the performance on ResNet-110.
Similar to what we have found with ResNet-56, HRank leads to an improvement in accuracy over the baseline model (94.23\% \emph{vs}. 93.50\%) with around 41.2\% FLOPs and 39.4\% parameters reduction.
Besides, compared with L1, with a slightly better complexity reduction, HRank again benefits from better accuracy performance (94.23\% \emph{vs}. 93.30\%).
Therefore, the rank can effectively reflect the relative importance of a filter and serve as a better intrinsic property.
Finally, in comparison with GAL-0.5, HRank greatly reduces the model complexity (58.2\% \emph{vs}. 48.5\% for FLOPs and 59.2\% \emph{vs}. 44.8\% for parameters) with only a small loss in accuracy of 0.14\%, while GAL-0.5 suffers a 0.95\% accuracy drop.
Moreover, we also maintain a similar accuracy with GAL-0.5 (92.65\% \emph{vs}. 92.55\%).
Obviously, HRank further accelerates the computation (68.6\% \emph{vs}. 48.5\%) and relieves the overload (68.7\% \emph{vs}. 44.8\%).
This indicates that HRank is especially suitable for pruning neural networks with residual blocks.

\textbf{DenseNet-40}.
Tab.\,\ref{densenet_cifar10} summarizes the experimental results on DenseNet.
We observe that HRank has the potential to remove more FLOPs.
Though Liu \emph{et al}. retains the accuracy of the baseline, it produces only 32.8\% FLOPs reductions.
In contrast, over 40.8\% of FLOPs are removed by HRank.
Compared with Zhao \emph{et al}. and GAL-0.05, HRank achieves higher accuracy and FLOPs reduction, though it removing fewer parameters.
Overall, HRank has the potential to better accelerate neural networks.
Hence, it can also be applied in networks with dense blocks.

\begin{table}[]
\scriptsize
\centering
\caption{Pruning results of ResNet-50 on ImageNet.}
\label{resnet_imagenet}
\begin{tabular}{ccccc}
\toprule
Model                                    &Top-1\%  &Top-5\%   &FLOPs     &Parameters\\
\midrule
ResNet-50 \cite{luo2017thinet}           &76.15    &92.87     &4.09B     &25.50M     \\
SSS-32 \cite{huang2018data}              &74.18    &91.91     &2.82B     &18.60M  \\
He \emph{et al}. \cite{he2017channel}    &72.30    &90.80     &2.73B     & -       \\
GAL-0.5 \cite{lin2019towards}            &71.95    &90.94     &2.33B     &21.20M  \\
\textbf{HRank}(Ours)                     &74.98   &92.33
&2.30B     &16.15M     \\
%
%
GDP-0.6 \cite{lin2018accelerating}       &71.19    &90.71     &1.88B     &-      \\
GDP-0.5 \cite{lin2018accelerating}       &69.58    &90.14     &1.57B     &-  \\
SSS-26 \cite{huang2018data}              &71.82    &90.79     &2.33B     &15.60M  \\
GAL-1 \cite{lin2019towards}              &69.88    &89.75    &1.58B     &14.67M  \\
GAL-0.5-joint \cite{lin2019towards}      &71.80    &90.82     &1.84B     &19.31M  \\
\textbf{HRank}(Ours)                     &71.98    &91.01  &1.55B     &13.77M \\
ThiNet-50 \cite{luo2017thinet}           &68.42    &88.30    &1.10B     &8.66M     \\
GAL-1-joint  \cite{lin2019towards}       &69.31    &89.12    &1.11B     &10.21M  \\
\textbf{HRank}(Ours)                     &69.10    &89.58
&0.98B     &8.27M       \\
\bottomrule
\end{tabular}
\vspace{-1.2em}
\end{table}

\subsubsection{Results on ImageNet}\label{results_on_imagenet}
We also conduct experiments for ResNet-50 on the challenging ImageNet dataset.
The results are shown in Tab.\,\ref{resnet_imagenet}.
Generally, HRank surpasses its counterparts in all aspects, including top-1 and top-5 accuracies, as well as FLOPs and parameters reduction.
More specifically, 1.78$\times$ FLOPs (2.30B \emph{vs}. 4.09B for ResNet-50) and 1.58$\times$ parameters (16.15M \emph{vs}. 25.50M of ResNet-50) are removed by HRank, while it still yields 74.98\% top-1 accuracy and 92.33\% top-5 accuracy, significantly better than GAL-0.5.
Moreover, HRank obtains 71.98 top-1 accuracy and 91.01 top-5 accuracy with 2.64$\times$ and 1.85$\times$ reductions of FLOPs and parameters, respectively.
Further, between ThiNet-50 and GAL-1-joint, we observe that GAL-1-joint advances in better accuracies, while ThiNet-50 shows more advantages in FLOPs and parameters reductions.
Nevertheless, HRank outperforms both GAL-1-joint and ThiNet-50.
Compared with GAL-1-joint, which gains 69.31\% top-1 and 89.12\% top-5 accuracies, HRank obtains 69.10\% top-1 and 89.58\% top-5 accuracies.
When comparing to ThiNet-50, better complexity reductions are observed (0.98B FLOPs \emph{vs}. 1.10B FLOPs for ThiNet-50 and 8.27M parameters \emph{vs}. 8.66M for ThiNet-50).
Hence, HRank also works well on complex datasets.

\subsection{Ablation Study}\label{ablation_study}
We conduct detailed ablation studies on variants of HRank and keeping a portion of filters frozen during fine-tuning.
For brevity, we report the results of the ResNet-56 with top-1 accuracy of 93.17\% in Tab.\,\ref{resnet_cifar10}.
Similar observations can be found in other networks and datasets.

\begin{figure}[!t]
\begin{center}
\includegraphics[height=0.55\linewidth]{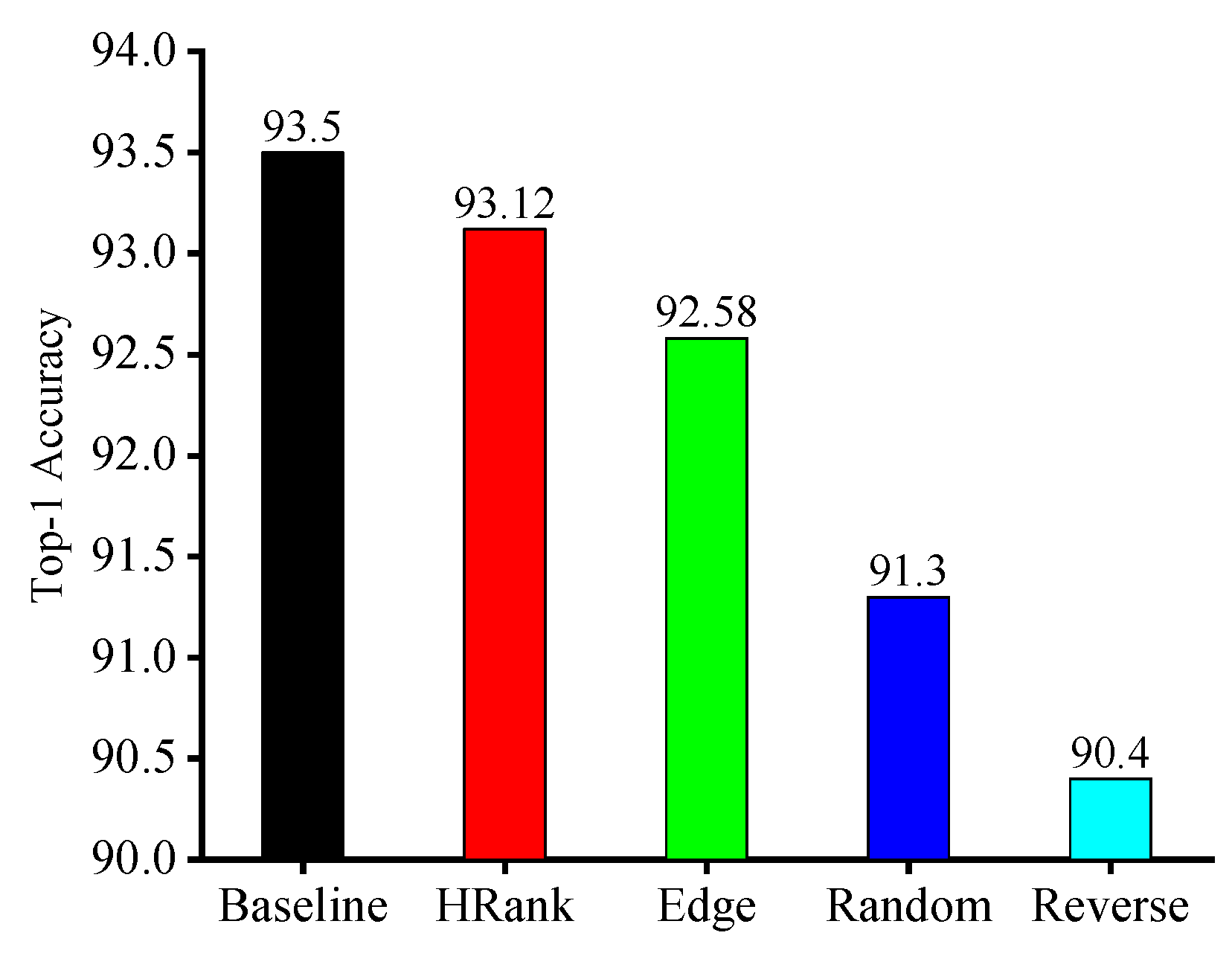}
\end{center}
\vspace{-1.0em}
\caption{\label{variant}Top-1 accuracy for variants of HRank.
}
\vspace{-1.2em}
\end{figure}

\textbf{Variants of HRank}.
Three variants are proposed to demonstrate the appropriateness of preserving filters with high-rank feature maps, including: (1) Edge: Filters generating both low- and high-rank feature maps are pruned, (2) Random: Filters are randomly pruned. (3) Reverse: Filters generating high-rank feature maps are pruned.
The pruning rate is set to be the same as that with 93.17\% top-1 accuracy of HRank in Tab.\,\ref{resnet_cifar10}.
We report the corresponding top-1 accuracy for each variant in Fig.\,\ref{variant}.
Among the variants, Edge shows the best performance.
To analyze, though part of filters with high-rank feature maps removed, it retains some of those with relatively high-rank feature maps, which contains rich information.
Besides, HRank obviously outperforms its variants and Reverse performs the worst, demonstrating that low-rank feature maps contain less information and the corresponding filters can be safely removed.

\begin{figure}[!t]
\begin{center}
\begin{minipage}[t]{0.85\linewidth}

\centerline{
\subfigure[$0\%$ of the filters are frozen, resulting in a top-1 precision of $93.17\%$.]{
\includegraphics[width=\linewidth]{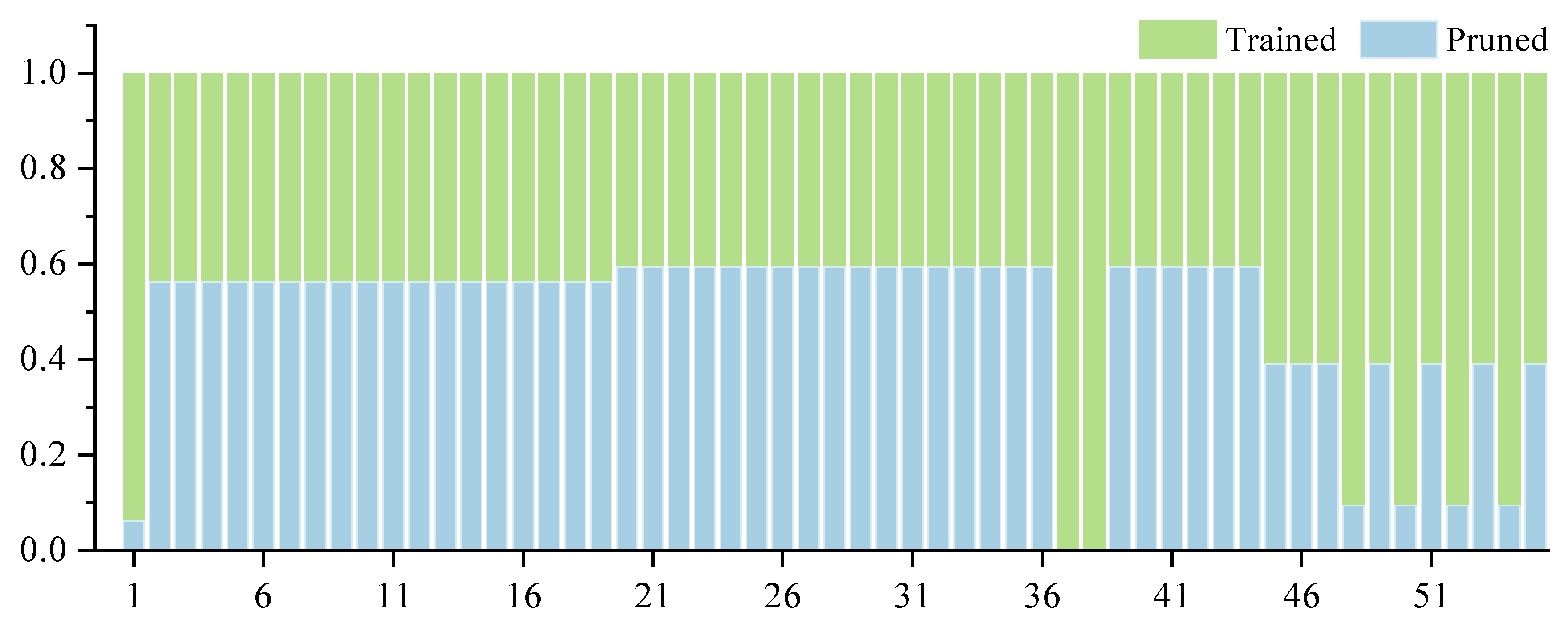}}
}

\centerline{
\subfigure[$15\%$ - $20\%$ of the filters are frozen, resulting in a top-1 precision of $93.13\%$.]{
\includegraphics[width=\linewidth]{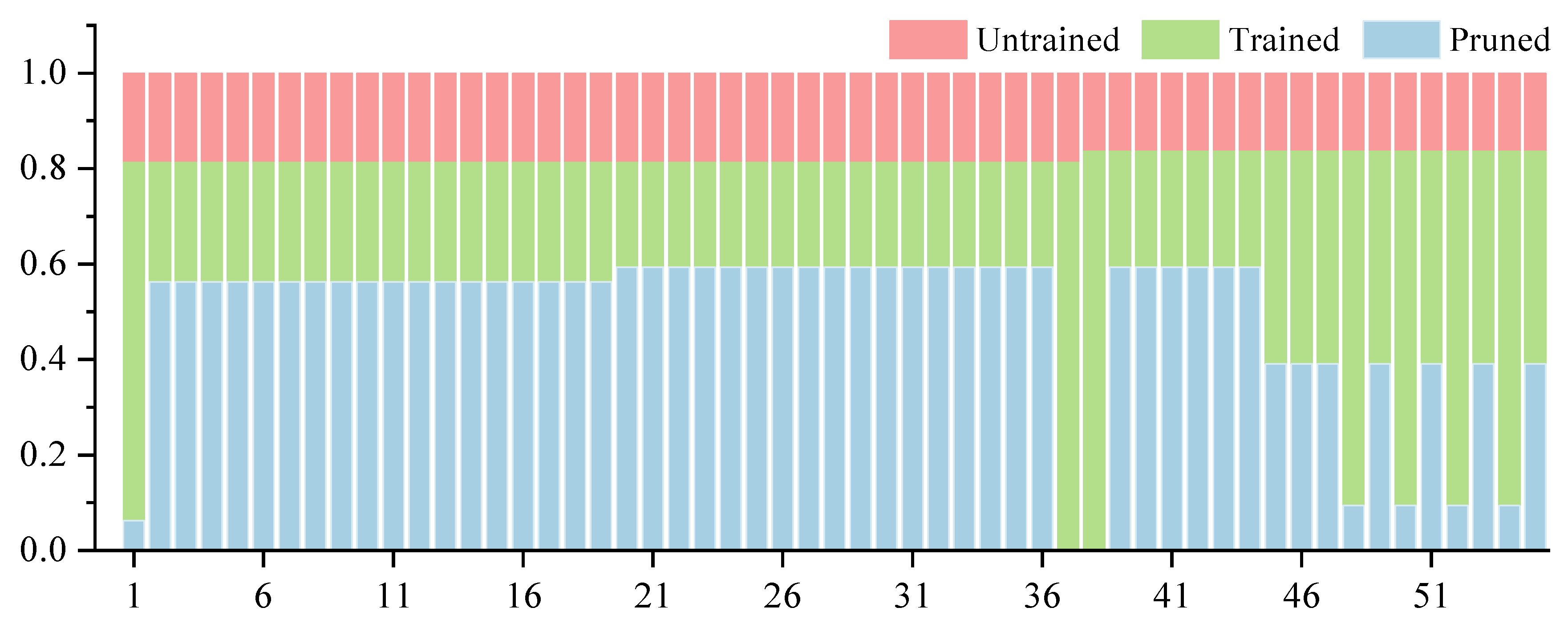}}
}

\centerline{
\subfigure[$20\%$ - $25\%$ of the filters are frozen, resulting in a top-1 precision of $93.01\%$.]{
\includegraphics[width=\linewidth]{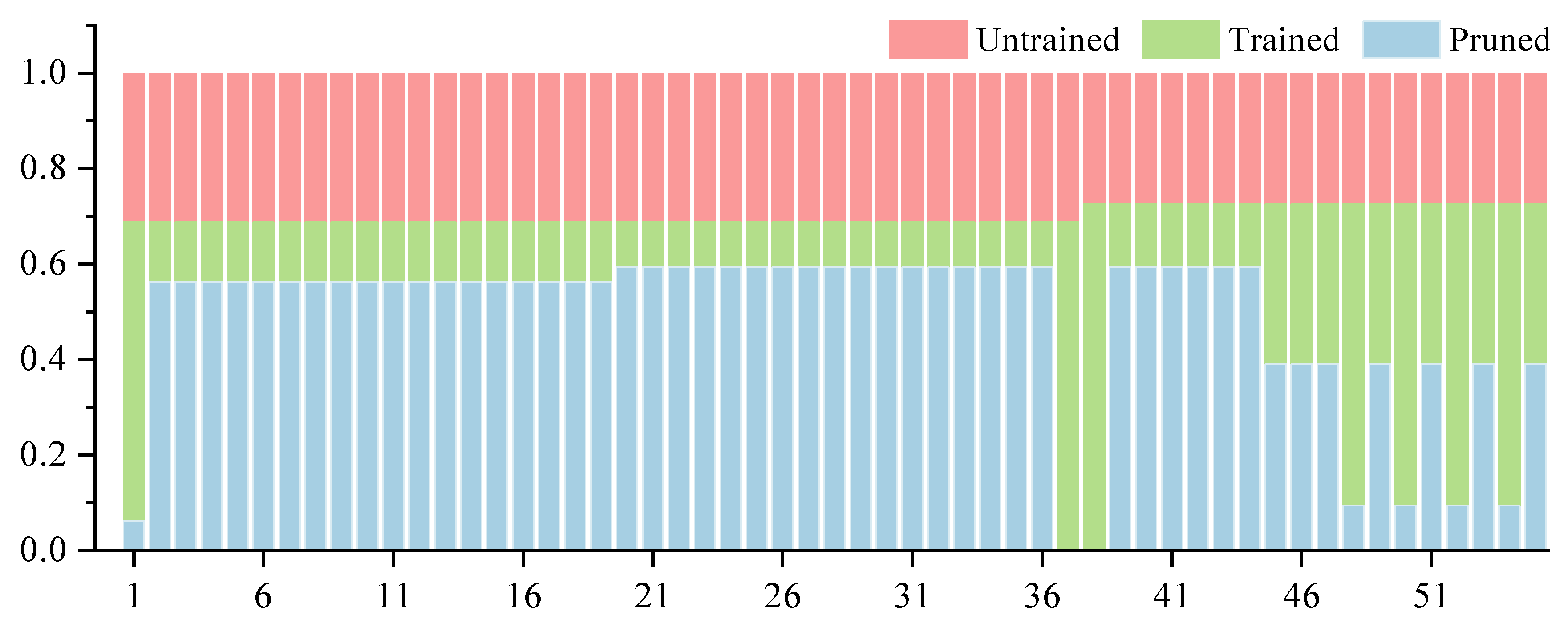}}
}
\end{minipage}

\end{center}
\vspace{-0.8em}
\caption{\label{noupdating}How freezing filter weights during fine-tuning affects the top-1 precision.
The x-axes denote the indices of convolutional layers.
The blue, green and red denote the percentage of pruned, trained and untrained filters, respectively.
}
\vspace{-1.5em}
\end{figure}

%
\textbf{Freezing Filters during Fine-tuning}.
We show that not updating filters with higher-rank feature maps does little damage to the model performance.
To that effect, we show the top-1 accuracy \emph{w.r.t.} different rates of untrained weights in Fig.\,\ref{noupdating}.
In Fig.\,\ref{noupdating}(a), all the reserved filters are involved in updating, which yields a top-1 accuracy of 93.17\% (this can be seen as the upper boundary for Fig.\,\ref{noupdating}(b) and Fig.\,\ref{noupdating}(c)).
In Fig.\,\ref{noupdating}(b), around 15\% - 20\% of filters are untrained, which returns a top-1 accuracy of 93.13\%, which is only slightly worse than the 93.17\% of Fig.\,\ref{noupdating}(a).
The effect of a larger percentage of untrained filters is displayed in Fig.\,\ref{noupdating}(c), consisting of around 20\% - 25\% of the overall weights.
The pruned model obtains a top-1 accuracy of 93.01\%.
Though with a 0.11\% drop in performance, it still outperforms the methods compared in Fig.\,\ref{resnet_cifar10}.
Fig.\,\ref{noupdating} well supports our claim that feature maps with high ranks contain more information.

\section{Conclusions}\label{conclusion}
In this paper, we present a novel filter pruning method called HRank, which determines the relative importance of filters by observing the rank of feature maps.
To that effect, we empirically demonstrate that the average rank of feature maps generated by a single filter are always the same.
Then, we mathematically prove that filters that generate lower-rank feature maps are less important and should be removed first, and vice versa, which is also verified by the experiments.
In addition, we propose to freeze a portion of the filters producing high-rank feature maps to further reduce the fine-tuning cost, showing little compromise to the model performance.
Extensive experiments on various modern CNNs demonstrate the effectiveness of HRank in reducing the computational complexity and model size.
In the future work, we will do more on the theoretical analysis on why the average rank of feature maps generated by a single filter are always the same.

\section{Acknowledge}
This work is supported by the Nature Science Foundation of China (No.U1705262, No.61772443, No.61572410, No.61802324 and No.61702136),
National Key R\&D Program (No.2017YFC0113000, and No.2016YFB1001503), and Nature Science Foundation of Fujian Province, China (No. 2017J01125 and No. 2018J01106).

{\small
\bibliographystyle{ieee_fullname}
\bibliography{main}

\begin{thebibliography}{10}\itemsep=-1pt

\bibitem{carreira2018learning}
Miguel~A Carreira-Perpin{\'a}n and Yerlan Idelbayev.
\newblock Learning-compression algorithms for neural net pruning.
\newblock In {\em Computer Vision and Pattern Recognition (CVPR)}, 2018.

\bibitem{chen2017deeplab}
Liang-Chieh Chen, George Papandreou, Iasonas Kokkinos, Kevin Murphy, and Alan~L
  Yuille.
\newblock Deeplab: Semantic image segmentation with deep convolutional nets,
  atrous convolution, and fully connected crfs.
\newblock {\em IEEE Transaction on Pattern Analysis and Machine Learning
  (TPAMI)}, 2017.

\bibitem{chen2015compressing}
Wenlin Chen, James T.~Wilson, Stephen Tyree, Killian Q.~Weinberger, and Yixin
  Chen.
\newblock Compressing neural networks with the hashing trick.
\newblock In {\em International Conference on Machine Learning (ICML)}, 2015.

\bibitem{cheng2018recent}
Jian Cheng, Pei-song Wang, Gang Li, Qing-hao Hu, and Han-qing Lu.
\newblock Recent advances in efficient computation of deep convolutional neural
  networks.
\newblock {\em Frontiers of Information Technology \& Electronic Engineering
  (FITEE)}, 2018.

\bibitem{denil2013predicting}
Misha Denil, Babak Shakibi, Laurent Dinh, Marc'Aurelio Ranzato, and Nando
  De~Freitas.
\newblock Predicting parameters in deep learning.
\newblock In {\em Neural Information Processing Systems (NeurIPS)}, 2013.

\bibitem{denton2014exploiting}
Emily~L Denton, Wojciech Zaremba, Joan Bruna, Yann LeCun, and Rob Fergus.
\newblock Exploiting linear structure within convolutional networks for
  efficient evaluation.
\newblock In {\em Neural Information Processing Systems (NeurIPS)}, 2014.

\bibitem{girshick2014rich}
Ross Girshick, Jeff Donahue, Trevor Darrell, and Jitendra Malik.
\newblock Rich feature hierarchies for accurate object detection and semantic
  segmentation.
\newblock In {\em Computer Vision and Pattern Recognition (CVPR)}, 2014.

\bibitem{han2016eie}
Song Han, Xingyu Liu, Huizi Mao, Jing Pu, Ardavan Pedram, Mark~A Horowitz, and
  William~J Dally.
\newblock Eie: efficient inference engine on compressed deep neural network.
\newblock In {\em International Conference on Computer Architecture (ISCA)},
  2016.

\bibitem{han2015deep}
Song Han, Huizi Mao, and William~J Dally.
\newblock Deep compression: Compressing deep neural networks with pruning,
  trained quantization and huffman coding.
\newblock In {\em International Conference of Learning Representation (ICLR)},
  2015.

\bibitem{han2015learning}
Song Han, Jeff Pool, John Tran, and William J~Dally.
\newblock Learning both weights and connections for efficient neural network.
\newblock In {\em Neural Information Processing Systems (NeurIPS)}, 2015.

\bibitem{he2016deep}
Kaiming He, Xiangyu Zhang, Shaoqing Ren, and Jian Sun.
\newblock Deep residual learning for image recognition.
\newblock In {\em Computer Vision and Pattern Recognition (CVPR)}, 2016.

\bibitem{he2019filter}
Yang He, Ping Liu, Ziwei Wang, Zhilan Hu, and Yi Yang.
\newblock Filter pruning via geometric median for deep convolutional neural
  networks acceleration.
\newblock In {\em Computer Vision and Pattern Recognition (CVPR)}, 2019.

\bibitem{he2017channel}
Yihui He, Xiangyu Zhang, and Jian Sun.
\newblock Channel pruning for accelerating very deep neural networks.
\newblock In {\em International Conference on Computer Vision (ICCV)}, 2017.

\bibitem{hu2016network}
Hengyuan Hu, Rui Peng, Yu-Wing Tai, and Chi-Keung Tang.
\newblock Network trimming: A data-driven neuron pruning approach towards
  efficient deep architectures.
\newblock {\em arXiv preprint arXiv:1607.03250}, 2016.

\bibitem{huang2017densely}
Gao Huang, Zhuang Liu, Laurens Van Der~Maaten, and Kilian~Q Weinberger.
\newblock Densely connected convolutional networks.
\newblock In {\em Computer Vision and Pattern Recognition (CVPR)}, 2017.

\bibitem{huang2018data}
Zehao Huang and Naiyan Wang.
\newblock Data-driven sparse structure selection for deep neural networks.
\newblock In {\em European Conference on Computer Vision (ECCV)}, 2018.

\bibitem{krizhevsky2009learning}
Alex Krizhevsky, Geoffrey Hinton, et~al.
\newblock Learning multiple layers of features from tiny images.
\newblock Technical report, Citeseer, 2009.

\bibitem{li2017pruning}
Hao Li, Asim Kadav, Igor Durdanovic, Hanan Samet, and Hans~Peter Graf.
\newblock Pruning filters for efficient convnets.
\newblock In {\em International Conference of Learning Representation (ICLR)},
  2017.

\bibitem{lin2020filter}
Mingbao Lin, Rongrong Ji, Shaojie Li, Qixiang Ye, Yonghong Tian, Jianzhuang
  Liu, and Qi Tian.
\newblock Filter sketch for network pruning.
\newblock {\em arXiv preprint arXiv:2001.08514}, 2020.

\bibitem{lin2020channel}
Mingbao Lin, Rongrong Ji, Yuxin Zhang, Baochang Zhang, Yongjian Wu, and
  Yonghong Tian.
\newblock Channel pruning via automatic structure search.
\newblock {\em arXiv preprint arXiv:2001.08565}, 2020.

\bibitem{lin2018holistic}
Shaohui Lin, Rongrong Ji, Chao Chen, Dacheng Tao, and Jiebo Luo.
\newblock Holistic cnn compression via low-rank decomposition with knowledge
  transfer.
\newblock {\em IEEE Transaction on Pattern Analysis and Machine Learning
  (TPAMI)}, 2018.

\bibitem{lin2018accelerating}
Shaohui Lin, Rongrong Ji, Yuchao Li, Yongjian Wu, Feiyue Huang, and Baochang
  Zhang.
\newblock Accelerating convolutional networks via global \& dynamic filter
  pruning.
\newblock In {\em International Joint Conference on Artificial Intelligence
  (IJCAI)}, 2018.

\bibitem{lin2019towards}
Shaohui Lin, Rongrong Ji, Chenqian Yan, Baochang Zhang, Liujuan Cao, Qixiang
  Ye, Feiyue Huang, and David Doermann.
\newblock Towards optimal structured cnn pruning via generative adversarial
  learning.
\newblock In {\em Computer Vision and Pattern Recognition (CVPR)}, 2019.

\bibitem{liu2017learning}
Zhuang Liu, Jianguo Li, Zhiqiang Shen, Gao Huang, Shoumeng Yan, and Changshui
  Zhang.
\newblock Learning efficient convolutional networks through network slimming.
\newblock In {\em International Conference on Computer Vision (ICCV)}, 2017.

\bibitem{long2015fully}
Jonathan Long, Evan Shelhamer, and Trevor Darrell.
\newblock Fully convolutional networks for semantic segmentation.
\newblock In {\em Computer Vision and Pattern Recognition (CVPR)}, 2015.

\bibitem{luo2017thinet}
Jian-Hao Luo, Jianxin Wu, and Weiyao Lin.
\newblock Thinet: A filter level pruning method for deep neural network
  compression.
\newblock In {\em Computer Vision and Pattern Recognition (CVPR)}, 2017.

\bibitem{molchanov2016pruning}
Pavlo Molchanov, Stephen Tyree, Tero Karras, Timo Aila, and Jan Kautz.
\newblock Pruning convolutional neural networks for resource efficient
  inference.
\newblock In {\em International Conference of Learning Representation (ICLR)},
  2016.

\bibitem{park2016faster}
Jongsoo Park, Sheng Li, Wei Wen, Ping Tak~Peter Tang, Hai Li, Yiran Chen, and
  Pradeep Dubey.
\newblock Faster cnns with direct sparse convolutions and guided pruning.
\newblock In {\em International Conference of Learning Representation (ICLR)},
  2016.

\bibitem{paszke2017automatic}
Adam Paszke, Sam Gross, Soumith Chintala, Gregory Chanan, Edward Yang, Zachary
  DeVito, Zeming Lin, Alban Desmaison, Luca Antiga, and Adam Lerer.
\newblock Automatic differentiation in pytorch.
\newblock In {\em Neural Information Processing Systems (NeurIPS)}, 2017.

\bibitem{ren2015faster}
Shaoqing Ren, Kaiming He, Ross Girshick, and Jian Sun.
\newblock Faster r-cnn: Towards real-time object detection with region proposal
  networks.
\newblock In {\em Neural Information Processing Systems (NeurIPS)}, 2015.

\bibitem{russakovsky2015imagenet}
Olga Russakovsky, Jia Deng, Hao Su, Jonathan Krause, Sanjeev Satheesh, Sean Ma,
  Zhiheng Huang, Andrej Karpathy, Aditya Khosla, Michael Bernstein, et~al.
\newblock Imagenet large scale visual recognition challenge.
\newblock {\em Internation Journal of Computer Vision (IJCV)}, 2015.

\bibitem{simonyan2014very}
Karen Simonyan and Andrew Zisserman.
\newblock Very deep convolutional networks for large-scale image recognition.
\newblock {\em arXiv preprint}, 2014.

\bibitem{szegedy2015going}
Christian Szegedy, Wei Liu, Yangqing Jia, Pierre Sermanet, Scott Reed, Dragomir
  Anguelov, Dumitru Erhan, Vincent Vanhoucke, and Andrew Rabinovich.
\newblock Going deeper with convolutions.
\newblock In {\em Computer Vision and Pattern Recognition (CVPR)}, 2015.

\bibitem{yu2018nisp}
Ruichi Yu, Ang Li, Chun-Fu Chen, Jui-Hsin Lai, Vlad~I Morariu, Xintong Han,
  Mingfei Gao, Ching-Yung Lin, and Larry~S Davis.
\newblock Nisp: Pruning networks using neuron importance score propagation.
\newblock In {\em Computer Vision and Pattern Recognition (CVPR)}, 2018.

\bibitem{zhang2015efficient}
Xiangyu Zhang, Jianhua Zou, Xiang Ming, Kaiming He, and Jian Sun.
\newblock Efficient and accurate approximations of nonlinear convolutional
  networks.
\newblock In {\em Computer Vision and Pattern Recognition (CVPR)}, 2015.

\bibitem{zhao2019variational}
Chenglong Zhao, Bingbing Ni, Jian Zhang, Qiwei Zhao, Wenjun Zhang, and Qi Tian.
\newblock Variational convolutional neural network pruning.
\newblock In {\em Computer Vision and Pattern Recognition (CVPR)}, 2019.

\bibitem{zhou2018revisiting}
Bolei Zhou, Yiyou Sun, David Bau, and Antonio Torralba.
\newblock Revisiting the importance of individual units in cnns via ablation.
\newblock {\em arXiv preprint arXiv:1806.02891}, 2018.

\end{thebibliography}
}

\end{document}